\documentclass[msc,ai,logo,twoside]{infthesis}

\usepackage{natbib}
\usepackage[]{algorithm2e}
\usepackage{graphicx}
\usepackage{amsmath}
\usepackage{spverbatim}
\DeclareMathOperator{\diag}{diag}
\DeclareMathOperator{\sigmoid}{sigmoid}
\DeclareMathOperator{\softmax}{softmax}

\title{Optimizing and Contrasting Recurrent Neural Network Architectures}
\author{Ben Krause}

\abstract{%
    Recurrent Neural Networks (RNNs) have long been recognized for their potential to model complex time series. However, it remains to be determined what optimization techniques and recurrent architectures can be used to best realize this potential. The experiments presented take a deep look into Hessian free optimization, a powerful second order optimization method that has shown promising results, but still does not enjoy widespread use. This algorithm was used to train to a number of RNN architectures including standard RNNs, long short-term memory, multiplicative RNNs, and stacked RNNs on the task of character prediction. The insights from these experiments led to the creation of a new multiplicative LSTM hybrid architecture that outperformed both LSTM and multiplicative RNNs. When tested on a larger scale, multiplicative LSTM achieved character level modelling results competitive with the state of the art for RNNs using very different methodology.

}

\begin{document}

\begin{preliminary}

\maketitle

\begin{acknowledgements}
I would like to thank Liang Lu and my supervisor Steve Renals for all their helpful suggestions on this project.

\end{acknowledgements}

\standarddeclaration

\tableofcontents

\end{preliminary}

\chapter{Introduction}

Feedforward neural networks are a powerful method for modelling static functions, that with enough hidden units can in theory model any function with arbitrary accuracy  \citep{Hornik-1989}. However, when modelling sequences, a simple feedforward neural network must have a fixed context of past inputs that it can use to make predictions. In order to use large contexts to make predictions, feed forward neural networks must use many inputs, and therefore must be much larger. A much more efficient and powerful way of using long time contexts is to give feedforward neural networks recursive connections, making them into recurrent neural networks (RNNs). RNNs have shown great potential to complex non-linear sequences that require long time range dependencies. However, there are many difficulties associated with training RNNs that make it hard to fully realize this potential. Additionally, as neural networks and datasets both become larger and highly parallelizable hardware such as GPUs improve, it is becoming increasingly important for training to become more efficient via parallelization than currently used methods. For these reasons, this study takes a deep look at training recurrent neural networks with Hessian free optimization \citep{Martens-2011}, which has shown potential to be both more effective and much more efficient via parallelization than commonly used training methods. There are many possible RNN architectures, some of which appear to be more expressive than others. Often, when one architecture outperforms another, it is difficult to know if the higher performing architecture is truly more expressive, or if the other architecture simply underperformed because the fitting algorithm was not powerful enough. For this reason, this study compares several different architectural ideas using Hessian free optimization for training, which is thought to be much less prone to under-fitting than other commonly used methods \citep{Martens-2010}. Some of these architectures have never been trained with Hessian free optimization, while others are novel architectures that have never been used, but likely would not be practical without the use of a powerful optimizer. For benchmarking these models, character level sequence modelling is used, a task that requires using highly non-linear transitions as well as long range dependencies to be successful. The goals of this study were to compare different ways of implementing Hessian free optimization as well as the expressiveness of several RNN architectures, and compare the best combinations to state of the art results for RNN character modelling.

\chapter{Recurrent neural networks}

Recurrent Neural Networks (RNNs) are a powerful class of sequence modellers that use recursive connections to store information about the past. The recursive nature of the hidden states of RNNs allows them to use a potentially unlimited context to make predictions. While the theoretical potential of recurrent neural networks to solve sequence modelling tasks has long been known, recently, improvements in optimization as well as computer hardware have allowed RNNs to achieve successful results at many supervised learning tasks. These achievements include state of the art results at speech recognition \citep{Graves-2013b}, language modelling \citep{Mikolov-2012c}, and handwriting recognition \citep{Graves-2009} among many others.

\section{Standard RNN}

The standard RNN architecture consists of one input vector $I$, one hidden vector $H$, and one output vector $O$. It contains 3 weights matrices of connections, input to hidden ($W_{hi}$), hidden to output ($W_{oh}$), and recurrent hidden to hidden ($W_{hh}$), and contains a vector of hidden state biases ($B_{h}$). Additionally, the output units typically have the $\softmax$ function applied to them to model a normalized probability distribution over the output units, and the hidden units have a non-linear squashing function such as the hyperbolic tangent ($\tanh$) applied. The RNN operates over a sequence of $T$ timesteps, and predicts $O(t)$ using $I(t)$ as well as passed context as shown in the equations below.
\begin{equation}
H(t) = \tanh(B_{h} + W_{hi}I(t) + W_{hh}H(t-1))
\end{equation}
\begin{equation}
O(t) = \softmax(W_{oh}H(t))
\end{equation}

Pseudocode for a recurrent neural network is also given in Algorithm 1.

    \begin{algorithm}[H]

 \For{t=1...T}{
  $H(t) \leftarrow B_{h}$\;
  $H(t) \leftarrow H(t) +W_{hi}I(t)$\;
  \If{$t>1$}{
   $H(t) \leftarrow H(t) + W_{hh}H(t-1)$\;

   }
  $H(t) \leftarrow \tanh(H(t))$\;
  $O(t) \leftarrow W_{oh}H(t)$\;
  $O(t) \leftarrow \softmax(O(t))$\;
 }
 \caption{Recurrent Neural Network}
\end{algorithm}
\hfill \break
An RNN can also be interpreted graphically with nodes to represent the state vectors and edges to represent weight matrices (Figure 2.1).

\begin{figure}

\includegraphics{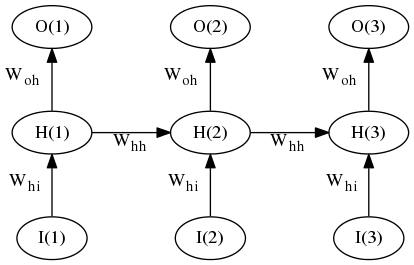}
\caption{A standard RNN}
\end{figure}

The performance of a recurrent neural network in modelling probabilistic outputs is evaluated by the cross entropy error between the RNNs outputs $O$, and the target outputs $\gamma$.
\begin{equation}
E(t) = - \gamma(t)\log(O(t))
\end{equation}

The partial derivatives of the cross entropy error with respect to the parameters of the network, $\frac{\partial E}{\partial \theta}$, also known as the gradient or direction of steepest descent, can be computed using the chain rule with the back propagation through time algorithm, which is given for a standard RNN with the following equations below \citep{Werbos-1990}.

\begin{equation}
\frac{\partial E}{\partial O(t))} = O(t) - \gamma(t)
\end{equation}
\begin{equation}
  \frac{\partial E}{\partial W_{oh}} = \sum_{t=1}^T \frac{\partial E}{\partial O(t)}H_{out}(t)^T
\end{equation}
\begin{equation}
\frac{\partial E}{\partial H_{out}(t)} =  W_{oh}^T \frac{\partial E}{\partial O(t)} + W_{hh}^T \frac{\partial E}{\partial H_{in}(t+1)}
\end{equation}
\begin{equation}
\frac{\partial E}{\partial H_{in}(t)} = \frac{\partial E}{\partial H_{out}(t)} \odot (1-H(t) \odot H(t))
\end{equation}
\begin{equation}
\frac{\partial E}{\partial W_{hh}} = \sum_{t=2}^T \frac{\partial E}{\partial H_{in}(t)}H(t-1)^T
\end{equation}
\begin{equation}
\frac{\partial E}{\partial W_{hi}} = \sum_{t=1}^T \frac{\partial E}{\partial H_{in}(t)}I(t)^T
\end{equation}

Note that the $\odot$ operator is the Hadamard product for element-wise matrix multiplication.

The most basic learning algorithm for an RNN would be to iteratively update the weights by a small negative factor of the gradient during training, an algorithm known as gradient descent. This can be done on single training cases of sequences segmented into fixed lengths, as is the case with stochastic gradient decent. Alternatively, training can become much more computationally efficient by using mini batches, in which the gradient is computed on several training cases simultaneously in parallel. It is also possible to use full batch gradient descent, in which the gradients are computed on the full dataset at once. While computationally efficient, this is generally impractical because the change to the sum of the error surfaces over all examples usually cannot be accurately approximated with a straight line, which is what gradient descent attempts to do. Furthermore, having stochasticity in training orders and updates provides training advantages, and this becomes no longer possible.

In general, training an RNN with gradient descent can result in a training difficulty known as the vanishing/exploding gradient problem \citep{Hochreiter-2001}, where the gradient tends to decay or explode exponentially as it is back-propagated through time. It can be seen why this problem arises by considering the matrix of derivatives of the hidden states at a given time point with respect to hidden states $n$ time steps in the past $\frac{\partial H_{out}(t)}{\partial H_{out}(t-n)}$.

\begin{equation}
\frac{\partial H_{out}(t)}{\partial H_{out}(t-n)} = \prod_{k=0}^{n-1} W_{hh}^{T} \diag(1-H(t-k)\odot H(t-k))
\end{equation}

For a large $n$, this matrix of derivatives will tend to either explode or decay exponentially because it is the product of many matrices. This result makes it difficult for RNNs to learn to use long-term dependencies in their predictions. When the gradient vanishes, the updates to the weights will not help with learning long time lags because this contribution to the gradient will be exponentially small, and when the gradient explodes learning becomes unstable. For this reason, more advanced architectures and/or learning algorithms are usually needed to train RNNs on difficult problems.

\section{Long short-term memory}

Long short-term memory (LSTM) is an RNN architecture designed to address some of the gradient based learning problems with the standard RNN architecture \citep{Hochreiter-1997}. It addresses this by having memory cells that use soft multiplicative gates to control information flow. In its original formulation, LSTM had input gates to control how much of the memory cell's total input is written to its internal state, and output gates to control how much of the memory cell's internal state is output. The memory cell had a self-recurrent weight of 1 preserving the internal state. Like in a standard recurrent neural network, the hidden state receives in inputs from the input layer and the previous hidden state
\begin{equation}
H_{in}(t) = W_{hi}I(t) + W_{hh}H_{out}(t-1)
\end{equation}

Sometimes, a non-linear squashing function is applied to $H_{in}(t)$. The input and output gates, $\omega$ and $\rho$, both also receive their own inputs, and typically use a $\sigmoid$ function to squash their input between 0 and 1, allowing them to act as soft, differentiable gates.

\begin{equation}
\omega(t) = \sigmoid(W_{\omega i}I(t) + W_{\omega h}H_{out}(t-1))
\end{equation}
\begin{equation}
\rho(t) = \sigmoid(W_{\rho i}I(t) + W_{\rho h}H_{out}(t-1)))
\end{equation}

The input gate controls how much of $H_{in}(t)$ is written to the LSTM cell's internal state, which has a self recurrent weight of 1. The input gate allows the cell to ignore irrelevant inputs.

\begin{equation}
  H_{state}(t) = H_{state}(t-1) + \omega(t) \odot H_{in}(t)\;
\end{equation}

The output gate then controls how much of the memory cells internal state is output, to be squashed with a non-linear function such as a hyperbolic tangent. The output gate allows the LSTM cell to keep information that is not relevant to the current output, but may be relevant later.
\begin{equation}
  H_{out}(t) = \tanh(H_{state}(t)\odot \rho(t));
\end{equation}

\begin{equation}
	O(t) = \softmax(W_{oh} H_{out}(t))
\end{equation}

The linear connections between successive internal states allow hidden states to be partially linear functions of their passed states, whereas in a standard RNN they become highly non-linear functions of their past states very quickly. Additionally, because the internal state has a self recurrent weight of 1, $\frac{\partial H_{state}(t)}{\partial H_{state}(t-1)}$ contains a term of 1, protecting against the vanishing gradient problem. In LSTM's original learning algorithm, exploding gradients were dealt with a type of derivative truncation in which $\frac{\partial E}{\partial H_{out}(t)}$ was approximated as $\frac{\partial E(t)}{\partial H_{out}(t)}$, thereby only allowing error to back-propagate though the LSTM's internal state. By including this truncation, it was assured that ${\frac{\partial H_{state}(t)}{\partial H_{state}(t-1)}}_{trunc} = 1$, and the approximated gradients would never vanish or explode. This also allowed for learning to occur completely online in real-time by making it possible to store ${\frac{\partial H_{state}(t)}{\partial\theta}}_{trunc}$ for all parameters of the network. This would be memory intensive to do for a standard recurrent neural network, but due to the approximations made to the gradient that force ${\frac{\partial H_{state}(t)}{\partial H_{state}(t-1)}}_{trunc} = 1$ , it holds that

\begin{equation}
{\frac{\partial H_{state}(t)}{\partial\theta}}_{trunc}= {\frac{\partial H_{state}(t-1)}{\partial\theta}}_{trunc} + \frac{\partial H_{state}(t)}{\partial\theta(t)}
\end{equation}

This recursion allows just one derivative value to be stored for every parameter in $\theta$, allowing for online approximations to the gradient to be computed in a memory efficient manner.

Forget gates, $\phi$, were later added to LSTM control how much of the memory cell's previous internal state is remembered, in place  of using a self-recurrent weight of 1 \citep{Gers-2000}. Forget gates receive input and are squashed non-linearly with a $\sigmoid$ in the same fashion as other gates. Their role in controlling information flow in the memory cell is given below.

\begin{equation}
H_{state}(t) = \phi(t) \odot H_{state}(t-1) + \omega(t) \odot H_{in(t)}\;
\end{equation}

Now $\frac{\partial H_{state}(t)}{\partial H_{state}(t-1)}$ includes a term equal to $\phi(t)$ instead of $1$, so the gradients once again can vanish, but will usually do so very slowly if $\phi$ is usually close to 1. Forget gates proved to be helpful for learning, and made LSTM more similar to standard RNNs. A diagram of information flow through an LSTM memory cell, now including forget gates, is presented in Figure 2.2. Note that in this diagram, passing through a square represents going through a multiplicative gate, and the edges simply represent the direction of information flow.

\begin{figure}[h]

\includegraphics[width=1.0\textwidth]{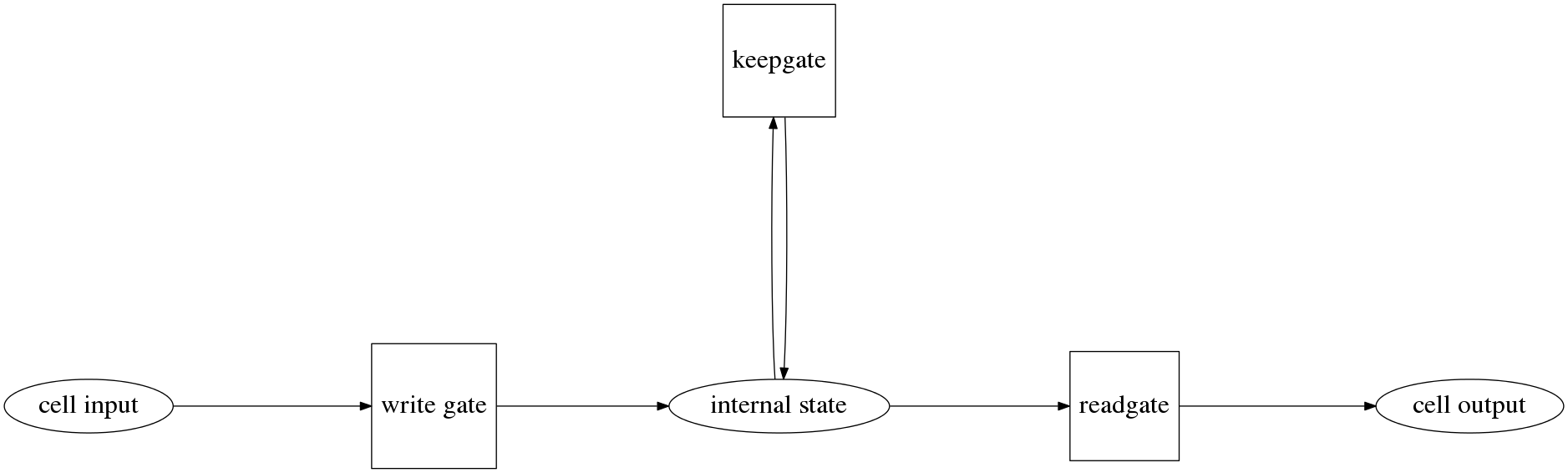}
\caption{LSTM cell}
\end{figure}

There are several variants of LSTM, and one of the simpler implementations based on the equations explained above is given in pseudocode below.

    \begin{algorithm}[H]

 \For{t=1...T}{

  $H_{in}(t) \leftarrow W_{hi}I(t)$\;
   $\omega(t) \leftarrow W_{\omega i}I(t)$\;
   $\phi(t) \leftarrow W_{\phi i}I(t)$\;
   $\rho(t) \leftarrow W_{\rho i}I(t)$\;

  \If{$t>1$}{
   $H_{in}(t) \leftarrow H_{in}(t) + W_{hh}H_{out}(t-1)$\;
     $\omega(t) \leftarrow \omega(t) + W_{\omega h}H_{out}(t-1)$\;
     $\phi(t) \leftarrow \phi(t) + W_{\phi h}H_{out}(t-1)$\;
     $\rho(t) \leftarrow \rho(t) + W_{\rho h}H_{out}(t-1)$\;

   }
   $\omega(t) \leftarrow \sigmoid(\omega(t))$\;
   $\phi(t) \leftarrow \sigmoid(\phi(t))$\;
   $\rho(t) \leftarrow \sigmoid(\rho(t))$\;
   $H_{state}(t) \leftarrow \omega(t) \odot H_{in}(t)$\;
   \If{$t>1$}{
   $H_{state}(t) \leftarrow H_{state}(t) + H_{state}(t-1)  \odot \phi(t)$
  }
  $H_{out}(t) \leftarrow \tanh(H_{state}(t)\odot \rho(t))$\;
  $O(t) \leftarrow W_{oh}H_{out}(t)$\;
  $O(t) \leftarrow \softmax(O(t))$\;
 }
 \caption{Long short term memory}
\end{algorithm}
\hfill \break

When LSTM was first invented, it was tested on a number of synthetic tasks that had previously been unsolvable by RNNs, designed to benchmark time-lag capabilities. An example of one of these tasks is the marked addition task \citep{Hochreiter-1997}, where the RNN receives a boolean and a linear input at every time step, and gives one linear output at the end of the sequence. The boolean input will be set to 1 twice, and 0 for every other time step. The RNN must learn to ignore the linear inputs when the boolean flag is set to 0, and at the final time-step must output the sum of the 2 marked linear inputs when the boolean flag was on. LSTM was able to solve this task even when the sequence was 100s of time steps long, meaning that it was able to learn to store the relevant inputs for hundreds of time-steps while ignoring irrelevant inputs. This was significant because it demonstrated that this architecture was able to use information from a very long time context. LSTM has also been very successful at natural tasks, achieving state of the art results at many supervising learning problems including handwriting recognition \citep{Graves-2009}, speech recognition \citep{Graves-2013b}, language modelling \citep{Zaremba-2014}, and RNN character prediction \citep{Graves-2013}.

\section{Stacked RNNs}

A commonly used strategy to increase the expressiveness of an RNN architecture is to stack RNNs sequentially, forming a sort of deep recurrent neural network hybrid \citep{Hermans-2013}, \citep{Pascanu-2013}. This allows for a greater deal of non-linear processing to occur between seeing input $I(t)$ and outputting output $O(t)$. It could also potentially allow for the RNN to store different time scales of information at each layer. The basic architecture of a stacked RNN is given in Figure 2.3.

 \begin{figure}[h]
 \includegraphics[width=1.0\textwidth]{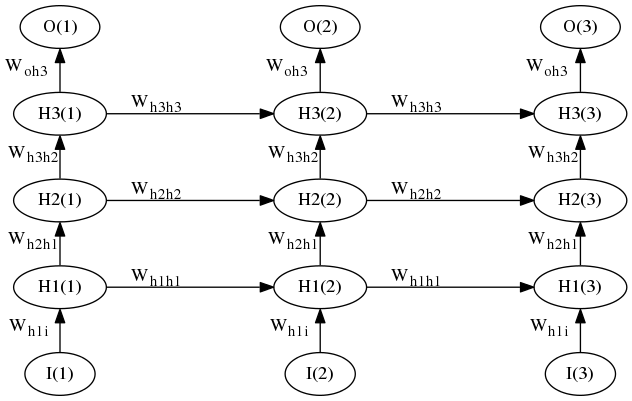}
\caption{A stacked RNN}
\end{figure}

One drawback of this architecture is that gradients now have to be back-propagated through feed forward layers as well as through time, magnifying exploding and vanishing gradient problems. Layer by layer training similar to what is commonly seen in deep neural networks has been helpful in training stacked RNNs \citep{Hermans-2013}. Another commonly used strategy to circumvent this is to give all the stacked RNN layers direct input and output connections \citep{Pascanu-2013}. This gives the RNN the flexibility of using varying degrees of non-linear processing when predicting output, intending to make training easier. The concept of stacked RNNs can be used with any RNN architecture, and is often found to improve results \citep{Hermans-2013}, \citep{Graves-2013}, \citep{Graves-2013b}.

\section{Multiplicative RNNs}

The multiplicative RNN is an architecture that was invented to allow varying transitions between hidden states depending on the input \citep{Sutskever-2011}. This is accomplished through a matrix factorization that allows the weights in the hidden transition to vary with a great degree depending on the input. The hidden to hidden weight matrix $W_{hh}$ in a standard RNN is replaced by a factorization with intermediate state M.
\begin{equation}
 W_{hh} = W_{hm} \diag(W_{mi}I(t))W_{mh}
\end{equation}

The effective hidden to hidden weight matrix that results from multiplying this out can be very different for each input I(t). For instance, the signs of effective hidden to hidden weights can be positive with some inputs and negative with others. The full equations for the architecture are given below.

\begin{equation}
H(t) = \tanh(B_{h} + W_{hi}I(t) +  W_{hm} \diag(W_{mi}I(t))W_{mh}H(t-1))
\end{equation}
\begin{equation}
O(t) = \softmax(W_{oh}H(t))
\end{equation}
 A diagram of a multiplicative RNN is given in Figure 2.4.

 \begin{figure}[h]
 \includegraphics[width=1.0\textwidth]{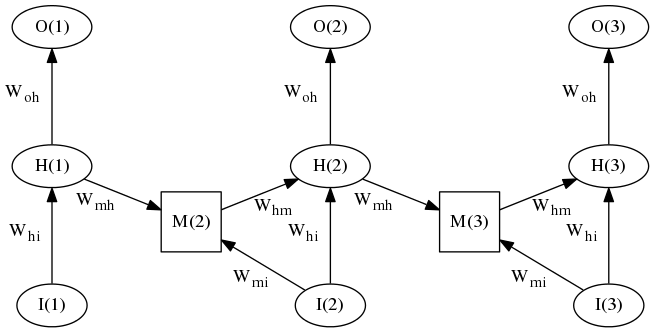}
\caption{A multiplicative RNN}
\end{figure}

Multiplicative RNNs are especially difficult to train with gradient descent because of the high degree of curvature in their error surface (see more on this in next chapter on optimization). For this reason, it is usually necessary to use 2nd order methods to train this architecture.

\section{Character prediction RNNs}

One of the more challenging sequence modelling tasks for an RNN is character prediction, which entails generating a probability distribution over a series of text characters. The RNN can start with the first character of a string $C_{1}$, and predict $P(C_{1})$, followed by $P(C_{2}|C_{1})$, $P(C_{3}|C_{2},C_{1})$, $P(C_{t+1}|C_{1}...C_{t})$, effectively generating a probability distribution over all possible sequences of characters. Typically, the RNN will receive $C_{t}$ as an input and try to predict $C_{t+1}$ at each time step. Performance is generally evaluated by the cross entropy of some unseen text in units of bits per character, computed as
\begin{equation}
\frac{1}{T} \sum_t (-\gamma(t)\log_{2}O(t))
\end{equation}

This is representative of the theoretical minimum number of bits per character that the sequence of text could be compressed to using the RNN as a probabilistic model. While modelling language at the word level is certainly easier because the model starts out with knowledge of the possible words, modelling language at the character level presents the potential to use sub-word segments to determine word meaning, as well as to correctly handle punctuation. Character level models are challenging to optimize, and push the limits of fitting algorithms, making them a good option for benchmarking advanced optimization methods.

Several experiments have been performed using RNN character level models. Two of the most widely benchmarked corpora have been the Penn Treebank corpus and several variations of the Wikipedia corpus. There have been a number of different RNN architectures tested on both of these corpora, and some of the main results are presented in Tables 2.1 and 2.2.

\begin{table}
\begin{center}
\begin{tabular}{ | l | l | l | l |} \hline Architecture & Test Error & Number of parameters \\ \hline RNN$^1$ & 1.41 & 420000 \\\hline  mRNN$^2$ & 1.41 & 4900000\\ \hline stacked LSTM$^3$  & 1.22 & 4300000 \\  \hline stacked RNN$^4$ & 1.41 &520000\\  \hline
\end{tabular}
\end{center}
\caption{RNN experiments on Penn Treebank corpus with test error given in bits/char, $^1$RNN \citep{Pascanu-2013}, $^2$mRNN \citep{Mikolov-2012b}, $^3$stacked LSTM using weight noise regularization and dynamic evaluation \citep{Graves-2013}, $^4$stacked RNN \citep{Pascanu-2013}}
\end{table}

\begin{table}[h]
\begin{center}
\begin{tabular}{ | l | l | l | l |} \hline Architecture & Test Error & Number of parameters \\ \hline mRNN$^1$ & 1.60 & 4900000 \\\hline mRNN$^2$ & 1.55 & 4900000 \\\hline  stacked RNN$^3$ & 1.47 & 9600000\\ \hline stacked LSTM$^4$  & 1.67 & 21300000 \\  \hline stacked LSTM$^5$ & 1.33 &21300000\\  \hline
\end{tabular}
\end{center}
\caption{RNN experiments on Wikipedia corpus, with test error given in bits/char, $^1$mRNN using split of 100 million character subset for both training and testing, with XML cut out of dataset \citep{Sutskever-2011}, $^2$mRNN using full 1.4 billion character training set 100 million character subset for testing, with XML cut out of dataset \citep{Sutskever-2011}, $^3$stacked RNN using full 1.4 billion character training set 100 million character subset for testing, with XML cut out of dataset \citep{Hermans-2013}, $^4$stacked LSTM using split of 100 million character subset for both training and testing, with XML included in dataset \citep{Graves-2013}, $^5$stacked LSTM using split of 100 million character subset for both training and testing, with XML included in dataset, using dynamic evaluation \citep{Graves-2013}}
\end{table}

\chapter{Optimization}

As stated in the previous chapter, there are certain difficulties in optimizing RNNs due to exponentially exploding and decaying gradients. A sensible solution to this problem is to either directly or indirectly use second order information in the optimization process. This will increase the step size in directions of low curvature, and decrease the step size in directions of high curvature, allowing for a much finer convergence. One reason that this is a sensible way to deal with the vanishing and exploding gradient problems is that the second derivatives are very likely to decay or explode at a similar rate to the first derivatives \citep{Martens-2010}, so the step sizes taken by a second order algorithm should naturally amplify decaying gradients and reduce exploding gradients appropriately.

\section{First order approximations}
Some simpler methods that indirectly use second order information involve slight modifications to first order algorithms to better control the step size in the direction of the gradient. One straightforward way of doing this is momentum, which is designed to accelerate convergence along valleys in the error surface by using a factor of previous updates as part of the current update. There are a number of ways to use momentum, but the classical way is given by the following equations \citep{Polyak-1964}
\begin{equation}
v_{t+1} = \mu v_{t} - \epsilon \nabla f(\theta_{t})
\end{equation}
\begin{equation}
\theta_{t+1} = \theta_{t} + v_{t+1}
\end{equation}
where $\mu$ is the momentum constant, $\epsilon$ is the learning rate, and $\nabla f(\theta_{t})$ is the gradient. When the gradient is small but consistent, meaning the second derivatives are low, using momentum will amplify updates to account for this. Another way to indirectly use second order information that has been found to work especially well for RNNs is gradient norm clipping. In this method, when the norm of the gradient exceeds a predefined threshold, the gradient is normalized so that the norm equals this threshold \citep{Mikolov-2012}. The idea behind this method is that if the gradient of an RNN is too high, it is likely the result of an exploding gradient and high second derivatives. Therefore in these cases, it makes sense to reduce the updates to account for the high curvature. First order methods that indirectly account for curvature can work well when the curvature is not too extreme. However, there are certain cases where modifying the step size in the direction of the gradient is not enough to find a good solution in a reasonable amount of time. To illustrate this, consider the multiplicative RNN (mRNN) architecture from the previous chapter, where the hidden to hidden transitions are given by
\begin{equation}
 W_{hh} = W_{hm} diag(W_{mi}I(t))W_{mh}
 \end{equation}
 With this hidden to hidden transition, many weights are being multiplied together, and a small change to one weight could greatly affect the gradients of many other weights in varying ways. This phenomenon is known as the pathological curvature problem of neural networks \citep{Martens-2010}, and it is especially severe in mRNNs. First order algorithms rely on the gradient direction being constant enough that it is possible to make reasonably significant step sizes in the direction of the gradient without the gradient completely changing. However, in an mRNN, even taking a very small step size in the direction of the gradient may cause the direction of steepest descent to completely change, making the original gradient no longer a viable search direction. While gradient descent could in theory train an mRNN with very small step sizes, this could take an impractically large amount of time. In order to deal with this, it is advisable to use second order methods that account for the full curvature matrix (or atleast some approximation to it) when computing updates. This allows the local error surface to be approximated with a quadratic bowl rather than a line, allowing for the approximation to be valid for a much larger radius when the curvature is high.

\section{Deriving Newton's method}

Second order methods are derived from the second order Taylor series local approximation of the error as a function of the change in parameters.

\begin{equation}
F(\theta + \Delta\theta) =  \frac{\Delta\theta^{T}H(F(\theta))\Delta\theta}{2} + \Delta\theta^{T}\nabla F(\theta) + F(\theta)
\end{equation}

where $\theta$ is a vector of all the parameters of the network, $F$ is the error as a function of these parameters, $\nabla F(\theta)$ is the gradient, and $H(F(\theta))$ is the Hessian matrix. Solving for the minimum by taking the derivative with respect to $\Delta\theta$ and setting this equal to 0 yields Equation 3.5.

\begin{equation}
\Delta\theta = -H(F(\theta))^{-1}\nabla F(\theta)
\end{equation}

The most straightforward optimization method that can be derived from this equation is Newton's method, which iteratively uses Equation 3.6 as an update rule until the function has converged.
\begin{equation}
\theta \leftarrow \theta - H(F(\theta))^{-1}(\nabla F(\theta))
\end{equation}

\section{Hessian free optimization}

While simple and straightforward, Newton's method would not be computationally feasible to train large neural networks. For a network with $N$ weights, storing the Hessian would be $O(N^{2})$ memory, and inverting it would be $O(N^{3})$ computation. Large neural networks often have millions of weights, making this training method impractical. A solution to this problem is to use an iterative truncated Newton's methods that can compute $-A^{-1}b$ for some matrix $A$ and some vector $b$ using only matrix vector products, a technique that has been well studied in the optimization community \citep{Nocedal-1999}, and recently introduced to machine learning. In the case of neural networks, methods have been developed to efficiently compute Hessian-vector products exactly with out ever having to store $H$  \citep{Pearlmutter-1994}. These methods generally would require $N$ iterations to compute $-A^{-1}b$ exactly when $A$ is a dense $NxN$ matrix and $b$ is an $Nx1$ vector. However, when $A$ is sparse or low rank, which is generally the case for the Hessian of neural networks, $-A^{-1}b$ can be approximated well in far fewer than $N$ steps.

\subsection{Conjugate gradient}
One of the most commonly used of such iterative methods is the conjugate gradient method. The conjugate gradient method can be contrasted to a simple steepest descent based method for solving a quadratic in the same form as the second order Taylor expansion
\begin{equation}
q(x) = \frac{x^TAx}{2} + b^Tx + c
\end{equation}

A steepest descent optimizer of this function would always step in the direction of the negative gradient, and would have an update rule of:
\begin{equation}
x \leftarrow x -(Ax+b)
\end{equation}

However, due to the off diagonal elements in A, the update directions will partially oppose previous updates, resulting in slower convergence. To prevent this, conjugate gradient ensures that each search direction $S_i$ is conjugate with respect to the previous search direction $S_{i-1}$ by adding an additional term to $S_i$. Conjugacy would be achieved by setting search directions to satisfy this condition:
\begin{equation}
S_{i}^{T} AS_{i-1}=0
\end{equation}

To enforce this, $S_{i}$ can be set equal to the negative gradient of the quadratic plus a scalar factor $\beta$ of the previous search direction that can be solved for to ensure conjugacy. The search direction will be in the form
\begin{equation}
S_i = -(Ax + b) + \beta S_{i-1}
\end{equation}

solving for $\beta$ to ensure conjugacy
\hfill \break
\begin{eqnarray}
-(Ax + b)^T A S_{i-1} + \beta S_{i-1}^{T} AS_{i-1}=0 \\
 \beta S_{i-1}^{T}AS_{i-1} = (Ax + b)^{T}AS_{i-1} \\
 \beta = \frac{(Ax + b)^{T} AS_{i-1}}{S_{i-1}^TAS_{i-1}}
\end{eqnarray}

\hfill \break

For each search direction S, the step size $\alpha$ that minimizes the quadratic Q must also be solved for.
\hfill \break
\begin{equation}
Q(x+\alpha S) = \frac{(x+\alpha S)^T A(x+\alpha S)}{2}  + b^{T}(x+\alpha S)
\end{equation}
Expanding this and eliminating terms without $\alpha$ yields the following term
\begin{center}
$\frac{(S^{T}AS)\alpha^2}{2} + (S^{T}Ax + b^{T}S)\alpha$
\end{center}
\hfill \break
Differentiating with respecting $\alpha$ and setting to 0 yields
\hfill \break
\begin{eqnarray}
(S^{T}AS)\alpha+ (S^{T}Ax + b^{T}S) = 0 \\
\alpha = \frac{-(S^{T}(Ax+b))}{(S^{T}AS)}
\end{eqnarray}

It can be shown that combining updates with these step sizes and directions will continually produce updates that are conjugate to all previous updates \citep{Shewchuk-1994}. The full conjugate gradient algorithm is given below.

\begin{algorithm}[H]

$x \leftarrow $ some initial guess \;

$S \leftarrow -(Ax + b)$\;
initial search direction equal to negative derivative of quadratic

 \For{i=1...N}{

	 $\alpha \leftarrow  \frac{-S^{T}(Ax + b)}{S^{T}AS}$\;
	 $x \leftarrow x+\alpha S$\;

	 $\beta \leftarrow \frac{(Ax+b)^{T}AS}{S^{T}AS}$\;
	 $S \leftarrow -(Ax+b) + \beta S$\;

 }
 \caption{Conjugate Gradient}
\end{algorithm}
\hfill \break

Note that the above derivation of conjugate gradient, \citep{Gibiansky-2014}, is somewhat different from, but mathematically equivalent to the form that is typically presented. The conjugate gradient method has several advantages over other iterative methods for solving linear systems, including being able to make use of a good initial guess, and not requiring memory storage from previous iterations.

Another important detail of applying conjugate gradient to neural networks is deciding when to terminate. The conjugate gradient algorithm in theory could take $N$ iterations to fully converge for a network with $N$ parameters, however, because the curvature matrix will almost always be sparse and/or low rank, most conjugate gradient methods applied to neural networks are limited to 50-200 iterations. Additionally, various heuristics can be used that terminate conjugate gradient early if progress in optimizing the quadratic approximation and or neural network error objective has nearly stopped. A commonly used heuristic is to terminate when the progress made in optimizing the quadratic after $k$ iterations is less than some tolerance $\epsilon$ per iteration \citep{Martens-2010}.
\begin{equation}
\frac{(q(i) - q(i-k))}{q(i)}<\epsilon k
\end{equation}

\subsection{Gauss-Newton matrix}

One potential limitation of the conjugate gradient method is that the curvature matrix must be positive semi-definite, or else the algorithm may converge to a maximum or a saddle point rather than a minimum. For this reason, an approximation to the Hessian matrix that only includes positive curvature known as the Gauss-Newton matrix is used in place of the Hessian. Note that from this point on, $H_{\sigma}$ will refer to the Hessian of the loss function, and any other references to H will refer to the hidden state of a neural network.
\begin{equation}
G = J(F(\theta))^T H_{\sigma}(F(\theta))J(F(\theta))
\end{equation}

Where $J(F(\theta))$ is the Jacobian matrix of derivatives, and $\sigma(F(\theta))$ is the loss function of F. $G$ will be positive semi-definite as long as a convex loss function is used. An efficient method for computing Gauss-Newton matrix vector products exactly for a neural network has also been derived \citep{Schraudolph-2002}, and can be achieved in 3 steps. To multiply the Gauss Newton Matrix G by some vector v, the first step is to compute Jv, which is equivalent to the directional derivative, or
\begin{equation}
J(F(\theta))v = \lim _{e\rightarrow 0} \frac{F(\theta+ev)}{e}
\end{equation}

A differential operator called the R-operator was derived to compute directional derivatives in neural networks \citep{Pearlmutter-1994}, and follows the rules of differential operators. It can be derived for a recurrent network (or any other differentiable network) to compute Jv using a few of these simple rules.

The notation $R(\cdot)$ is shorthand for the notation $R_v(\cdot)$, and refers to the directional rate of change of $\cdot$ in the direction v. For propagating weight matrices from layer $H_{1out}$ to layer $H_{2in}$, using the product rule for matrices:

\begin{equation}
R(H_{2in}) =  WR(H_{1out}) + R(W)H_{1out}
\end{equation}

Note that R(W) is just the directional change in W for which Jv is being computed; R(W) is set to the portion of v which refers to W. Similarly, in the case where $H_3 = H_2 \odot H_1$,
\begin{equation}
R(H_3) = R(H_2) \odot H_1 + H_2 \odot R(H_1)
\end{equation}

In the case where $H_3= H_2+H_1$, then $R(H_3) = R(H_2) + R(H_1)$ This is also the case when adding a fixed parameter bias, however in that case $R(B_h)$ would just be set equal to the part of v referring to $B_h$. lastly, in the case where a function is applied to $H_{in}$ to yield $H_{out} $
\begin{equation}
R(H_{out}) = R(H_{in}) \odot \frac{\partial H_{out}}{\partial H_{in}}
\end{equation}

Using these basic rules of differentiation, the R-forward algorithm can be derived for an RNN. The full algorithm is given below.

\begin{algorithm}[H]

This algorithm must applied after (or concurrently with) the forward pass algorithm, because H(t) values are needed

	use same $\theta$ (weight matrices and biases) as forward pass

 \For{t=1...T}{
  $R(H(t)) \leftarrow R(B_{h})$\;
  $R(H(t)) \leftarrow R(H(t)) +R(W_{hi})I(t)$\;
  \If{$t>1$}{
   $R(H(t)) \leftarrow R(H(t)) + R(W_{hh})H(t-1)$\;

      $R(H(t)) \leftarrow R(H(t)) + W_{hh}R(H(t-1))$\;

   }
  $R(H(t)) \leftarrow R(H(t))\odot(1-H(t)\odot H(t))$\;
  $R(O(t)) \leftarrow R(W_{oh})H(t)$\;
  $R(O(t)) \leftarrow  R(O(t)) + W_{oh}R(H(t))$\;

 }
 \caption{R-forward pass to compute directional derivatives of an RNN}
\end{algorithm}
\hfill \break

The algorithm above for computing Jv is approximately the same computational cost as 2 forward passes. The next step is to multiply this result by the Hessian of the loss function. In the case of an MSE loss function, this Hessian is simply the identity matrix. In the case of the cross-entropy error with softmax output units, this Hessian at each time step t is
\begin{equation}
H_\sigma = \diag(O(t))-O(t)O(t)^{T}
\end{equation}

In either case this Hessian-vector product can be computed very quickly compared to the other steps in computing Gv, so long as it is computed with element wise multiplication instead of actually computing the matrix vector products, which would be very inefficient for a matrix than can be expressed with just one vector.

The last step, multiplication by $J^{T}$, is exactly the same as applying the back propagation through time algorithm (see Chapter 2), but instead using
\hfill \break
$\frac{\partial E}{\partial (O(t))} = H_{\sigma}R(O(t))$, in place of
$\frac{\partial E}{\partial (O(t))} = O(t) - \gamma(t)$
\hfill \break

While being forced to use the Gauss-Newton matrix in place of the Hessian could be viewed as a drawback of conjugate gradient, even other iterative solvers that are capable of using the full Hessian have been found to perform better with the Gauss-Newton matrix approximation \citep{Vinyals-2011}. This may be because negative contributions to the curvature could tend have higher third derivatives and lead to a less stable approximation. Additionally, obtaining Gauss-Newton matrix vector products for neural networks requires only about half the computation of obtaining Hessian-vector products \citep{Schraudolph-2002}.

\subsection{Damping}

In order for this second order method to work well for training neural networks, one additional modification needs to be made. The second order Taylor series approximation is only a local approximation, however conjugate gradient will converge to the global minimum of the approximation, which is usually far outside the region for which it the approximation is accurate. A technique called ``damping'' is used to prevent this from happening. One simple way to perform damping, known as Tikhonov damping, that works reasonably well for feedforward neural networks, is to add a constant $\lambda$ to the diagonal elements of the Gauss-Newton matrix \citep{Martens-2010}. By conservatively overestimating the curvature, larger updates can be prevented. This method of damping turns out to be insufficient for training recurrent neural networks \citep{Martens-2011}, most likely because small changes in the weights can still sometimes lead to large changes in the network dynamics. Another type of damping called ``structural damping'' was created to cope with this problem \citep{Martens-2011}.

\subsubsection{Structural damping}

Structural damping creates a penalty on updates based on the changes they make to the hidden state dynamics, rather than the magnitude of the update. It uses a second quadratic approximation in which the error is the change to hidden states, which is added to the initial quadratic optimization objective to form a new objective. The structural damping loss function can be formulated as
\begin{equation}
E(H) = \mu (H(\theta) - H(\theta'))^2
\end{equation}

where $H(\theta)$ is the hidden state values for a given set of network parameters, and $\mu$ is a meta parameter that controls the weighting of this loss function. Note that in previous experiments that used both Tikhonov damping and structural damping, the loss function of structural damping used a weight of $\mu\lambda$. However, this thesis contains experiments that use structural damping only, so here $\mu$ will refer to the weight of structural damping regardless of if $\lambda$ is also used. The change in the hidden state is also approximated with the second order Taylor series. The first order term in the Taylor series is irrelevant in this case because the error of this objective is 0 with $\theta' = \theta$. The second order term, the curvature in the change in hidden states, can be accounted for with the Gauss-Newton matrix of the change in hidden states $G_s$. Computing the curvature-vector products $G_{s}v$ can be done by applying the R-forward algorithm and back-propagating the resulting values of $R(H(t))$. Since the steps for doing this are quite similar to the steps for computing the original Gauss-Newton matrix-vector products, the weighted sum of $G$ and $\mu G_{s}$ can be computed with just one slight modification to the backward pass step \citep{Martens-2011}.

\begin{equation}
\frac{\partial R}{\partial O(t)} = H_{\sigma}R(O(t))
\end{equation}
\begin{equation}
  \frac{\partial R}{\partial W_{oh}} = \sum_{t=1}^T \frac{\partial R}{\partial O(t)} H_{out}(t)^T
\end{equation}
\begin{equation}
\frac{\partial R}{\partial H_{out}(t)} =  W_{oh}^T \frac{\partial R}{\partial O(t)} + W_{hh}^T \frac{\partial R}{\partial H_{in}(t+1)}
\end{equation}

The $\mu R(H(t))$ term accounts for structural damping
\begin{equation}
\frac{\partial R}{\partial H_{in}(t)} = \frac{\partial R}{\partial H_{out}(t)} \odot (1-H(t) \odot H(t)) + \mu R(H(t))
\end{equation}
\begin{equation}
\frac{\partial R}{\partial W_{hh}} = \sum_{t=2}^T \frac{\partial R}{\partial H_{in}(t)}H(t-1)^T
\end{equation}
\begin{equation}
\frac{\partial R}{\partial W_{hi}} = \sum_{t=1}^T \frac{\partial R}{\partial H_{in}(t)}I(t)^T
\end{equation}

Note that in the above equations, there is an implicit multiplication of $H_{\sigma}(H(t)) R(H(T))$, however since MSE loss is being used for structural damping, $H_{\sigma}(H(t))$ is just the identity matrix. With this slight modification to the Gauss Newton matrix, changes to the RNNs parameter vector which cause the quadratic to be inaccurate can kept under control. When using Hessian free optimization with structural damping, standard RNNs first became capable of solving synthetic time-lag problems that were previously only solvable by LSTM \citep{Martens-2011}.

\subsubsection{Line-search damping}

Another perspective on damping is that the ideal damping parameters may vary a great deal for different points in training, or even for  different updates of the same conjugate gradient run \citep{Sutskever-2013b}. While techniques exist for adjusting parameters based on heuristic information, it is unlikely that damping parameters can be adjusted quickly enough to become optimal for each search direction. In the conjugate gradient method, the conjugacy of the search directions ensures that each direction optimizes the quadratic independently. This justifies only stepping in these search directions to the degree which they minimize the loss of the neural network. In this alternative method of damping, separate updates are made to the quadratic and the neural network parameters. The total quadratic update $x_{q}$ is optimized exactly as before with each step size $\alpha$ computed to minimize the quadratic approximation along search direction $S_{i}$, for n iterations of conjugate gradient
\begin{equation}
x_{q} = \sum_{i=1}^{n} \alpha{_i} S_{i}
\end{equation}
The update to the neural network parameters $x_{f}$ uses the same search direction, but adjusts the step sizes $\epsilon$ to greedily minimize the loss function with a line search.
\begin{equation}
x_{f} = \sum_{i=1}^{n} \epsilon{_i}\alpha{_i} S_{i}
\end{equation}

For the line-search to find each $\epsilon_i$, it makes sense to start with $\epsilon=1$, and backtrack, since updating the full direction along the optimal search direction for the approximated quadratic will almost always be an overestimate. This line-search can be done cheaply because it does not need to be exact.

 \begin{algorithm}[H]

	define decay constant $\tau$\;
	define maxIterations \;
	$F \leftarrow $ evaluative function at $\alpha S_i$\;
	\For{i=1..maxIterations}{

	$F_2  \leftarrow $ evaluative function at $\tau \epsilon \alpha S_i$\;
	\If{$F_2 < F$}{

		$F \leftarrow F_2$\;
		$\epsilon \leftarrow \tau \epsilon$\;
		}
	\Else{
		return\;

	}
	}

  \caption{Backtracking line-search used with line-search damping}
\end{algorithm}
\hfill \break

This damping method is rarely used, but a similar variant was found to outperform structural damping on an RNN controller task \citep{Sutskever-2013b}.

\subsection{Overview of Hessian free optimization}

Now that the steps of the Hessian free optimization algorithm have been derived, a high level view of the algorithm is presented.

\begin{algorithm}[H]

 initialize $\theta$

 \For{i=1...number of training iterations}{

 	randomly select training examples $n_j$ to compute gradient with

    Compute forward pass
    Compute 	$\nabla F(\theta)$ using back propagation through time

    select $n_c$, subset of $n_j$, to compute curvature matrix-vector products on

    Run conjugate gradient algorithm using 	$\nabla F(\theta)$ as $b$, and computing $AS_i$ products using steps described to compute dampened Gauss Newton matrix-vector products

	set v to update from conjugate gradient algorithm

    $\theta \leftarrow \theta + v$

    }

 \caption{High level view of Hessian free optimization}
\end{algorithm}

\subsection{Batch sizes}

One big advantage of using Hessian free optimization over first order methods is that it can make use of much larger batch sizes during training. While first order methods must rely on many small updates to converge, accounting for curvature allows Hessian free optimization to make much larger updates, making it suitable for larger batch sizes. Larger batch sizes allows for much greater efficiency on GPUs or multi-threaded CPUs via parallelization.

\subsubsection{Parallelization}
There are two levels of parallelization that can be used when training neural networks. Parallelization at the matrix multiplication level can occur when propagating state vectors during the forward and backward passes of a neural network. For instance in the forward pass of an RNN with $h$ units and $n$ training examples when
\hfill \break
$H(t) \leftarrow H(t) + W_{hh}H(t-1)$
\hfill \break

In the step where $W_{hh}$ is multiplied by $H(t-1)$, an $hxh$ matrix is being multiplied by a $hxn$ matrix. The larger $n$ is, the greater amount of local parallelization that can be performed, especially on a GPU. It is also possible to divide up computations on completely separate threads. In the case of computing gradients and Gv products, the results on each thread computing on a subset of examples can be summed to yield the total values.

When attempting to train on long sequences with large batch sizes, the memory requirements may become prohibitive, especially on GPUs, which often have less RAM and do not manage memory as efficiently. To compute the backward passes of RNNs, all previous hidden states values are needed. Computing the hidden states in the reverse direction by inferring $H(t-1)$ from $H(t)$ and $I(t)$ results in an exponential loss of precision, and is therefore completely impractical. Storing all the hidden state values has a memory requirement of $O(hnT)$, where $h$ is the hidden state dimensionality, $n$ is the number of training examples, and $T$ is the sequence length. Fortunately, there is a useful work around that reduces this to $2(hn \sqrt{T}) = O(hn \sqrt{T})$ that results in doubling the computation of the forward pass with the following steps \citep{Martens-2012}

  \hfill \break
 	1. compute forward pass, store the hidden state vector at intervals of $\sqrt{T}$
 	\hfill \break
 	2. compute backward pass, recompute forward pass for every interval $(i-1) \sqrt{T}..i \sqrt{T}$ resulting in the storage of $\sqrt{T}$  additional hidden vectors at any given time.
\hfill \break

Note that it is possible to be slightly more efficient by storing at shorter intervals at the end of the sequence, and throwing them away during the backward pass, however this is still $O(hn \sqrt{T})$.

While it may seem inefficient to have to effectively compute the forward pass twice, the time lost is typically much less than the time gained by being able to multiply much larger matrices on a GPU at once (for both forward and backward passes) as opposed to having to compute over training examples sequentially with smaller matrices.

\subsubsection{Choosing batch sizes for gradient and curvature computations}

Choosing $n_j$, the batch size for computing the gradient, and $n_c$, the batch size for computing curvature matrix vector products are important decisions for balancing efficiency and successful training in Hessian free optimization. If the batches are too small, parallelization may be less efficient and the optimizer may over-fit mini-batches and slow down training. If the mini-batches are too large, the neural network could lose the advantages of stochasticity in the examples used, which can help prevent settling to local optimum. In general, $n_j$ is atleast several times larger than $n_c$, and the curvature batch is a subset of the gradient batch. The reasoning for $n_j$ being larger is that the gradient only needs to be computed once for every run of conjugate gradient, and each run of conjugate generally has many iterations, which all require curvature matrix vector products to be computed. It is sensible to get a more accurate approximation to the gradient because it must be computed far less frequently. In terms of general batch sizes, the optimal number may be task specific. For character level modelling, past work has shown that computing the gradient on a few million characters and the curvature on a few hundred thousand characters works well \citep{Sutskever-2011}. This would entail splitting up the characters into $n$ sequences of length $T$ in such a way that $nT$ was in these ranges. In general, some experimentation may be needed to determine optimal batch sizes for any given task, but previous work can give good indications of approximately what to start with.

\chapter{Preliminary experiments}

\section{Overview of methods}
The initial set of experiments carried out were intended as a benchmarked comparison between different methodologies of RNN character level modelling. These experiments were run on a subset of the Penn Treebank corpus, containing the first 2.8 million characters for training, and approximately 200,000 for validation and 200,000 for testing. These experiments used the Penn Treebank corpus in its raw form, with no preprocessing, and numbers and rare words included$^1$. The smaller subset was used to allow models to be trained in a reasonable amount of time on CPUs. The RNN architectures were designed to all have around 215-220 thousand trainable parameters, including weights and biases. In all runs, training was stopped when the validation error stopped improving. These experiments were carried out using custom-written MATLAB code. All experiments were run on a machine with 8 Quad core AMD Opteron 6328 processors (3.2 GHz each), using 10 threads when paralellization was possible (all Hessian free optimization experiments). Multithreading was performed using MATLAB's Parallel Computing Toolbox.

The first set of experiments compared the original structural damping method for RNNs and mRNNs with line-search damping. The second set of experiments compared LSTM trained with Hessian free optimization (structural damping) to the original 1st order online learning algorithm for LSTM. The next set of experiments compared a stacked mRNN trained with Hessian free optimization (structural damping) to earlier results. The last experiment in this chapter tested out a novel multiplicative LSTM hybrid architecture, trained with Hessian free optimization (structural damping).

For the structural damping, $\mu$ was adjusted according the following formulas \citep{Martens-2011}

let $f(i)$ be the error function being minimized on the $i$th iteration of conjugate gradient, and $q(i)$ quadratic approximation function being minimized on the $i$th iteration of conjugate gradient

\begin{equation}
p=\frac{(f(i)-f(0))}{(q(i)-q(0))}
\end{equation}

When $p<0.25$, $\mu \leftarrow 2\mu /3$, and when $p>0.75$, $\mu \leftarrow 3\mu /2$,

Several conditions were set for conjugate gradient termination. Firstly, to stop conjugate gradient when progress was no longer being made, the formula from the previous chapter was used with $k = 10$, and $\epsilon = 0.0005$
\begin{equation}
\frac{(q(i) - q(i-k))}{q(i)}<\epsilon k
\end{equation}

To terminate conjugate gradient when the quadratic approximation was no longer accurate, termination occurred if $\mu(i) >= 3\mu_0$. For line-search damping only, conjugate gradient was also terminated when the line-search failed to make progress more than 5 times. The line-search does not make progress when it decays the search direction for a predefined maximum number of iterations (set to 10) and fails to improve upon the error of the network from before the update. Lastly, partially for efficiency purposes, conjugate gradient was limited to a maximum of 100 iterations. At the end of each conjugate gradient run, a backtracking line-search was used to find the step size that minimized the error to the training set.

\hfill \break

$^1$The capital letters 'T', 'U' and 'V', as well as a few commonly used symbols such as the dollar sign were left out by mistake, and all mapped to a single character. This was fixed for the full-scale experiments (next chapter), and should not have had any significant impact on the comparisons in this chapter which all had these characters missing.

\section{Damping}

Structural damping, which penalizes updates by how much change to the hidden state they are expected to result in, has proven to be a viable option for second order training of RNNs. However, an alternative method that was named ``Augmented Hessian free optimization'' was suggested to be more robust \citep{Sutskever-2013b}. However, only one experiment was provided to demonstrate this, in which an RNN controller trained with this method resulted in slower initial convergence, but a slightly better final outcome than structural damping. Because of the limited research into this method, experiments comparing a similar approach were carried out here. The exact implementation details of this method were spared in the original paper \citep{Sutskever-2013b}, so it is difficult to know if the implementation here (presented in Chapter 3.3 as line-search damping) is identical, but at the very least it is inspired by the same insights.

Experiments were carried out comparing structural damping to line-search damping for standard RNNs and multiplicative RNNs. The standard RNN had 400 hidden units, and the mRNN had 280 hidden units. A sparse initialization of the hidden to hidden matrix of the standard RNN was used, designed to be similar to what was used in past RNN experiments with Hessian free optimization \citep{Martens-2011}. Parameters in $W_{hh}$ were initialized to $0$ with a probability of $0.9$, and initialized independently with a normal distribution with mean $0$ and standard deviation $0.1$ with a probability of $0.1$. All input and output weights were initialized with a normal distribution with a mean of 0 and standard deviation of $0.1$. For the multiplicative RNN, all weights were initialized with a mean of $0$ and a standard deviation of $0.05$. For each iteration of training, the gradient was computed on the full dataset split into sequences of length 200, and a random selection of sequences totalling one fourth of the dataset was used to compute Gauss-Newton matrix-vector products for conjugate gradient runs.

For structural damping, the meta-parameter $\mu$ was set to $0.01$ for standard RNNs and $0.3$ for mRNNs.

The results of the experiments are given in Table 4.1.

\begin{table}[h]
\begin{center}
\begin{tabular}{ | l | l | l | l | l |} \hline Architecture/damping & Train Error & Val Error & Test Error & Number of parameters \\ \hline RNN/structural & 1.80 & 2.04 & 1.99 & 216400 \\\hline  RNN/line-search & 1.78 & 2.03 & 1.97 & 216400\\ \hline mRNN/structural & 1.69 & 1.90 & 1.87 & 215880\\ \hline mRNN/line-search & 1.72 & 1.88 & 1.85 & 215880\\ \hline
\end{tabular}
\end{center}
\caption{Results of standard RNN and mRNN trained with structural damping and line-search damping in bits/char}
\end{table}

The use of the line-search damping resulted in a slightly better generalization error in both cases, however, it also took slightly longer to fit. Line-search damping also made less progress in the early iterations, which is consistent with a similar past approach \citep{Sutskever-2013b}.

 \begin{figure}[h]
 \includegraphics[width=1\textwidth]{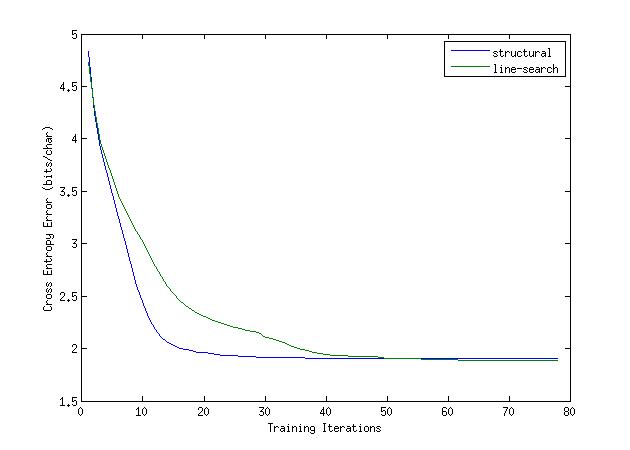}
\caption{Validation error for line-search damping and structural damping during training}
\end{figure}

Another result demonstrated here is that the multiplicative RNN performs significantly better than the standard RNN regardless of damping. This is consistent with past results, and it is thought that the multiplicatives RNN's ability to alter the transition matrix as a function of the input gives it an advantage for fitting character level models.

Overall these results showed that line-search damping is a viable alternative to structural damping for training RNNs with Hessian free optimization. They also provided baseline values for other experiments.

\section{LSTM}

Until recently, LSTM had been far more successful than other RNNs at solving problems that required long range dependencies. When Hessian free optimization was first applied to RNNs, it was questioned whether the LSTM architecture was actually more expressive for natural tasks, or had just achieved success because it had been easier to train without advanced optimization methods \citep{Martens-2011}. The authors of this study compared an LSTM trained with stochastic gradient descent with classical momentum to a standard RNN with a similar number of parameters trained with Hessian free optimization, and found that the standard RNN achieved a superior performance on a variety of natural tasks including bouncing balls video prediction, speech prediction, and music prediction \citep{Martens-2011}. However, with this learning algorithm, the LSTM may have had some training problems due to exploding gradients, which LSTMs are still vulnerable to. The most successful LSTM learning algorithms typically use derivative clipping to protect against this \citep{Graves-2013}. While the authors of the paper may not have known about gradient clipping methods at the time, it may have provided a fairer comparison to also use Hessian free optimization to train the LSTMs. To my knowledge, no published result has ever used Hessian free optimization to train LSTM. Doing so would allow for a fairer comparison with standard RNNs also trained with the same algorithm, and also could provide an alternative way to train LSTMs on large datasets with increased efficiency via paralellization.

The only step that needs to change to train LSTM with Hessian free optimization is the algorithm for multiplying Gauss-Newton matrix-vector products. Recalling from the previous chapter, a Gauss Newton matrix-vector product $Gv$ is expressed as
\begin{equation}
Gv = J(F(\theta))^T H_{\sigma}(F(\theta))J(F(\theta))v
\end{equation}

Computing $J(F(\theta))v$ requires applying the R-forward algorithm to an LSTM, which can be derived easily using the basic rules given in Chapter 3.3. The equations for the R-forward algorithm for LSTM are given in Section 1 of the appendix. Multiplying the result of the R-forward algorithm by $H_{\sigma}(F(\theta))$ is exactly the same as described in the previous chapter because the nature of LSTM output units is the same. Multiplying by $J(F(\theta))^T$ is simply the back-propagation through time algorithm for LSTM applied to the result of the first 2 steps. It can be described by the equations in section 2 of the appendix, which can also be used to compute the gradient of LSTM.

Two experiments were carried out on identical LSTM architectures, one that used Hessian free optimization (structural damping, $\mu = 0.01$), and another that used the original LSTM online learning algorithm described in Section 2.2 on LSTM. The original LSTM online learning algorithm gives updates similar to stochastic gradient descent, and prevents exploding gradients through its error truncation. It is still used for online dynamic adaptation of character level models \citep{Graves-2013}. All weights in both experiments were initialized randomly with a mean of 0 and a standard deviation of 0.1. The results of these experiments are given in table 4.2, along with the standard RNN baseline from the previous section.

\begin{table}[h]
\begin{center}
\begin{tabular}{ | l | l | l | l | l |} \hline Architecture & Train Error & Val Error & Test Error & Number of parameters \\ \hline RNN & 1.80 & 2.04 & 1.99 & 216400 \\\hline  LSTM Hessian free & 1.75 & 1.93 & 1.88 & 220350\\ \hline LSTM online & 1.81 & 2.03 & 1.98 & 220350\\  \hline
\end{tabular}
\end{center}
\caption{Results of LSTM trained with Hessian free optimization and LSTM trained with original online algorithm, compared with previous result of standard RNN trained with Hessian free optimization in bits/char}
\end{table}

Note that the total training time for LSTM trained with Hessian free optimization was an order of magnitude less because of the utilization of multithreading, which is not possible for the online LSTM learning algorithm. Interestingly, the LSTM trained with Hessian free optimization achieved a superior performance to both the original LSTM learning algorithm as well as the standard RNN trained with Hessian free optimization. Hessian free optimization is able to deal with exploding gradients in such a way that it can still use the full gradient, which likely gives it an advantage over the online LSTM learning algorithm with error truncation. LSTM's superior performance over the standard RNN suggests that atleast for this specific task of character prediction, LSTM is likely to be a more expressive architecture than the standard RNN. There are couple of possible reasons for this. The most obvious one is that the gated internal states of the LSTM may help it use a longer time context in making predictions, which is very important when modelling language at a character level. Another possibility is that similarly to the mRNN, the LSTM is able to have more non-linearity in its transitions depending on the input character due to its input gates and forget gates.

\section{Stacked mRNNs}

Stacking RNNs is a sensible way to increase the expressiveness of any RNN architecture. Stacked RNN variants have been used in several of the most successful RNN character prediction results \citep{Graves-2013}, \citep{Hermans-2013}.
However, previous papers with mRNNs have never attempted to stack them. While mRNNs are able to have highly non-linear transitions that in some sense already make them deep, making them deep in a completely different way could certainly increase their expressiveness. For this reason, a stacked mRNN was trained on this data set for comparison with the shallow mRNN. To make trainer easier, all layers in the stacked mRNN were given direct input and output connections. The stacked mRNN had 3 total layers, and its units were distributed from first layer to last layer as 150-130-110.

For the notation, $l$ refers to the layer, $W_{mli}$ is the input connection matrix to the multiplicative intermediate state for layer $l$, $W_{mlh}$ is the hidden connection matrix to the multiplicative intermediate state for layer $l$, $W_{hli}$ is the input to hidden connection matrix for hidden layer l, $W_{olh}$ is the output connection matrix from hidden layer $l$, $W_{hlm}$ is the multiplicative to hidden connection matrix for layer $l$, $W_{hlh}$ is the input from hidden layer $l-1$, which only exists when $l>1$, and $B_{hl}$ is the bias terms for hidden layer $l$. $L$ is the total number of stacked hidden layers.

 \begin{equation}
M_{l}(t) = W_{mli}I(t) \odot (W_{mlh}H_l(t-1))
   \end{equation}

 \begin{equation}
H_l(t) = \tanh(B_{hl} + W_{hli}I(t) + W_{hlm}M_{l}(t) + W_{hlh}H_{l-1}(t)
   \end{equation}
\begin{equation}
O(t) = \softmax( \sum_{l=1}^{L} W_{olh}H_l(t))
\end{equation}

The stacked RNN was trained with Hessian free optimization (structural damping, $\mu = 0.3$). All weights were initialized randomly with a mean of 0 and a standard deviation of 0.05. The results of the stacked mRNN are given in table 4.3, with the original mRNN from section 4.1 also given for comparison.
\begin{table}[h]
\begin{center}
\begin{tabular}{ | l | l | l | l | l |} \hline Architecture & Train Error & Val Error & Test Error & Number of parameters \\  \hline mRNN & 1.69 & 1.90 & 1.87 & 215880\\ \hline Stacked mRNN & 1.74 & 1.90 & 1.87 & 219090\\ \hline
\end{tabular}
\end{center}
\caption{Comparison of results of stacked mRNN to previous result of mRNN in bits/char}
\end{table}

Interestingly, the performance of a stacked mRNN does not seem to differ much from the original mRNN. While stacking architectures often leads to some improvement, one past experiment that used stacked RNNs at character prediction on this same corpus  experienced almost no improvement over standard RNNs, despite the stacked RNN having a significantly greater number of parameters \citep{Pascanu-2013}. It is possible that the Penn Treebank corpus does not have long enough passages or enough variation in words for stacked mRNNs to provide any significant advantage. It may also be the case that stacked mRNNs would be easier to regularize, however this was not tested here.

\section{Multiplicative LSTM}

Using multiplicative hidden to hidden weights, and using LSTM cells, both individually resulted in improvements in performance. Of course, these two ideas are not mutually exclusive, and may help RNNs perform better for different reasons. The factorized intermediate states of mRNNs allows them to be very flexible in their hidden to hidden transition depending on the input. While the input gates and forget gates give LSTMs some flexibility in their transition, they have nowhere near the flexibility of mRNNs. LSTMs on the other hand may have an advantage in storing information undisturbed for longer periods of time. This could be a useful ability for mRNNs to have, seeing as information could be more difficult to store for long periods of time over highly complex transitions. For these reasons, a novel multiplicative LSTM (mLSTM) hybrid architecture was created and tested on this data set. mLSTM has a factorized intermediate state M with the same dimensionality as the hidden state. This intermediate state is the same as in a regular mRNN.

$M(t) = (W_{mh}H_{out}(t-1)) \odot (W_{mi}I(t))$

All hidden units and gate units then receive input from this multiplicative intermediate

\begin{equation}
\omega(t) = \sigmoid(W_{\omega i}I(t) + W_{\omega m}M(t))
\end{equation}
\begin{equation}
\phi(t) = \sigmoid(W_{\phi i}I(t) + W_{\phi m}M(t))
\end{equation}
\begin{equation}
\rho(t) = \sigmoid(W_{\rho i}I(t) + W_{\rho m}M(t))
\end{equation}
\begin{equation}
H_{in}(t) = W_{hi}I(t) + W_{h m}M(t)
\end{equation}

The LSTM cell then operates exactly the same way as described in the LSTM chapter. The full pseudocode for an mLSTM is provided below.

 \begin{algorithm}[H]

 let $\phi$ be the forget gate vector, $\omega$ be the input gate vector, and $\rho$ be the output gate vector
 \For{t=1...T}{

  $H_{in}(t) \leftarrow W_{hi}I(t)$\;
   $\omega(t) \leftarrow W_{\omega i}I(t)$\;
   $\phi(t) \leftarrow W_{\phi i}I(t)$\;
   $\rho(t) \leftarrow W_{\rho i}I(t)$\;

  \If{$t>1$}{

		$M(t) = (W_{mh}H_{out}(t-1)) \odot (W_{mi}I(t))$
  }
   $H_{in}(t) \leftarrow H_{in}(t) + W_{hm}M(t))$\;
     $\omega(t) \leftarrow \omega(t) + W_{\omega m}M(t))$\;
     $\phi(t) \leftarrow \phi(t) + W_{\phi m}M(t))$\;
     $\rho(t) \leftarrow \rho(t) + W_{\rho m}M(t))$\;

   $\omega(t) \leftarrow \sigmoid(\omega(t))$\;
   $\phi(t) \leftarrow \sigmoid(\phi(t))$\;
   $\rho(t) \leftarrow \sigmoid(\rho(t))$\;
   $H_{state}(t) \leftarrow \omega(t) \odot H_{in}(t)$\;
   \If{$t>1$}{
   $H_{state}(t) \leftarrow H_{state}(t) + H_{state}(t-1)  \odot \phi(t)$
  }
  $H_{out}(t) \leftarrow \tanh(H_{state}(t)\odot \rho(t))$\;
  $O(t) \leftarrow W_{oh}H_{out}(t)$\;
  $O(t) \leftarrow \softmax(O(t))$\;
 }
 \caption{mLSTM}
\end{algorithm}
\hfill \break
Like the other architectures described, The R-forward pass can be used to compute $Jv$. This is derived for mLSTM in Section 3 of the appendix.

Back propagation can be applied to compute the backward pass in computing $Gv$, as well as to compute the gradients. This is derived for mLSTM in Section 4 of the appendix.

mLSTM was tested on this same dataset, also trained with Hessian free optimization, using structural damping with $\mu=0.1$ (a run comparing $\mu=0.1$ and $\mu=0.3$ found that $\mu=0.1$ gave a validation error of $1.86$ bits/char and $\mu=0.3$ gave $1.87$ bits/char). All weights were initialized randomly with a mean of 0 and a standard deviation of 0.1. The results compared to earlier results are presented in Table 4.4.

\begin{table}[h]
\begin{center}
\begin{tabular}{ | l | l | l | l | l |} \hline Architecture & Train Error & Val Error & Test Error & Number of parameters \\  \hline mRNN & 1.69 & 1.90 & 1.87 & 215880\\ \hline LSTM Hessian free & 1.75 & 1.93 & 1.88 & 220350\\ \hline mLSTM & 1.69 & 1.86 & 1.82 & 215900\\ \hline
\end{tabular}
\end{center}
\caption{Comparison of results of mLSTM hybrid architecture in bits/char to previous results of an mRNN and LSTM}
\end{table}

mLSTM was able to improve upon the results of LSTMs and mRNNs alone, suggesting that there are some advantages in combining these architectures. It seems that the combination of being able to have highly input dependant transitions, while also being able to store protected hidden states that can persist these transitions works reasonably well for this task.

\section{Overview of results}

Several general conclusions can be made from the results of this chapter. The line-search damping provides an alternative to structural damping that leads to slower initial converge but potentially slightly better generalization. LSTM cells seemed to provide an advantage over standard RNN units, and this advantage was realized most fully when the LSTM was also trained with Hessian free optimization. Additionally, multiplicative hidden weights provided an advantage over the standard hidden weights. Combining LSTM cells with multiplicative hidden weights into a novel multiplicative LSTM architecture led to the best overall results. An overview of all experiments from this chapter is presented in Table 4.5.

\begin{table}[h]
\begin{center}
\begin{tabular}{ | l | l | l | l | l |} \hline Architecture & Train Error & Val Error & Test Error & Number of parameters \\ \hline RNN/structural & 1.80 & 2.04 & 1.99 & 216400 \\\hline  RNN/line-search & 1.78 & 2.03 & 1.97 & 216400\\ \hline mRNN/structural & 1.69 & 1.90 & 1.87 & 215880\\ \hline mRNN/line-search & 1.72 & 1.88 & 1.85 & 215880\\ \hline LSTM online & 1.81 & 2.03 & 1.98 & 220350\\  \hline LSTM Hessian free & 1.75 & 1.93 & 1.88 & 220350\\ \hline Stacked mRNN & 1.74 & 1.90 & 1.87 & 219090\\ \hline mLSTM & 1.69 & 1.86 & 1.82 & 215900\\ \hline
\end{tabular}
\end{center}
\caption{Results of all preliminary experiments together in bits/char}
\end{table}

\chapter{Full-scale experiments}

\section{Penn Treebank (full set) experiments}

\subsection{Methods}

Moving forward, mLSTM, the most successful model from the earlier experiments, was tested on the full Penn Treebank dataset. The Penn Treebank corpus is divided into subsections labelled as 00 through 24. As per standard procedure, sections 00-20 were used for training, 21-22 were used for validation, and 23-24 were used for testing. The total training set consists of approximately 5 million characters. Like in the previous experiments, the raw unprocessed corpus was used. Two models were with trained with different numbers of parameters. In both cases, the models were trained with Hessian free optimization with structural damping ($\mu=0.1$). All weights were initialized randomly with a mean of 0 and standard deviation of 0.1. Random sets of 2.16 million characters were used to compute gradients, and random subsets of 216 thousand characters were used for conjugate gradient runs. All training sets were split into sequences of length 300. The experiments were run on a 3 GPU machine containing 2 Geforce GTX 980s(4GB) and 1 Nvidia Tesla 40k(12GB). The GPUs were utilized using MATLAB's Parallel Computing Toolbox. The other specific details for the training procedure were essentially the same as in the previous chapter.

\subsection{Results}

\begin{table}[h]
\begin{center}
\begin{tabular}{ | l | l | l | l | l |} \hline Architecture & Train Error & Val Error & Test Error & Number of parameters \\  \hline mLSTM medium & 1.60 & 1.74 & 1.70 & 603000\\ \hline mLSTM large & 1.40 & 1.76 & 1.72 & 3608000\\ \hline
\end{tabular}
\end{center}
\caption{Results of multiplicative LSTM on raw Penn Treebank corpus in bits/char}
\end{table}

The results of this experiment are presented in Table 5.1. While these experiments were originally intended to be comparable to RNN results in literature, it turned out that the significant amount of preprocessing that been applied to the Penn Treebank corpus used for past RNN character level models had made the task much easier \citep{Mikolov-2012b}. This preprocessing limited the vocabulary to 10,000 unique words, mapping all rare words such as names to the same token, mapped all numbers to the capital letter N, lower-cased all words, and removed punctuation. All of these preprocessing steps make the text considerably more predictable, especially because names and numbers occur frequently in the corpus, and are among the hardest segments to predict. Not surprisingly, the results in RNN literature indicate a lower cross entropy than what was obtained here. For instance, an mRNN trained with Hessian free optimization that had a test set cross entropy error of 1.41 bits/char on the preprocessed corpus \citep{Mikolov-2012b}. It certainly seems plausible that the model tested here would be able to do atleast this well on the preprocessed corpus, but follow up experiments using the preprocessed corpus would be needed to directly compare this model to models in the literature.

In general, these results do indicate that over-fitting was a significant problem. Strong evidence for this is that the larger model performed worse despite having a much greater fitting capability. Part of the reason for this is that the corpus is relatively small, so the model may over-fit to specific numbers, names, and other rare words in the training data. Since Hessian free optimization is such a powerful fitting method and mLSTM is such an expressive model, regularization methods will need to be developed to make the most of this combined system.

\subsection{Samples from model}
Another way to benchmark character level models is to see what text is generated when sampling from the model. Samples can be taken from an RNN by starting with a short string of text to give the RNN context, and then having the RNN predict the probability distribution over the next character. The next character is then chosen probabilistically according to the RNN output, and the chosen character is used as the next input into the RNN. By doing this repeatedly, one can generate samples of text. In the Penn Treebank corpus, new passages start with the string ``START.'', so for these samples, the string ``START.'' was given to the RNN, and the RNN generated samples with this initial context.
1000 character samples were taken from the model with 600k parameters and the model with 3.6 million parameters. The results are given below.
\hfill \break
\hfill \break
600k parameter mLSTM

\begin{spverbatim}
START.

Rubos himself, and depens to Bull added $207 million, for a purchase ill 40 agreement, while 1986, softer value of sales in Pa," and 255 square control by and ever had been able to truly did also weldy every approval of the California auto maker has affect," Connie Hifts.
In a proposed 7.05%
The company said he had a various center and she desperately not think that the "certainly: "And next research -- of corporators and a small weaker, Mr. Lubers Telesies Xter's supporting." Warner could stand being override supports in few years to bought we're natured or long into our power at 2 de the companies in the Stock Exc
\end{spverbatim}

\hfill \break
3.6 million parameter mLSTM

\begin{spverbatim}
START.
Then, had little vegetable for delegation.

Kenneth Browns", a while financial income tax rates in a bad looks like the clinzents, Amulan Ore., lately.
Both Jacob Markei.

Japanese company, currently has still took position from the House attorney's airlines such as they called a former general boosting operations.

The Businesship, kept it."

If the rufflies customers we feel in a $36.3 million common stock.

.START

The Internal Webster mach.
Barry Japan continues that will offer prosecutions.

But the matter's loss was

.START

Think was accepted any symptory or themselves to possibly don't include the two days to the program by most oil and defenious new indexwest about the syndicate purse."

SuM R. Hatcherate G," says Michael Miller, N.G.

"I agreed would negotiate joined

\end{spverbatim}

The text sometimes makes sense for a few words, but rarely has any meaning beyond that. Since this is a character level model, it also sometimes invents words, some of which are more plausible than others. Because this is a probabilistic sample, the output will sometimes be a character that was fairly unlikely under the model, but happened to be chosen by chance. This makes it difficult for all words to be real words, or for the text to have any meaning beyond a short context. Overall though, it is clear from this that the model did learn a fair amount of words, sentence structure, punctuation, and capitalization considering that it had no built in knowledge of any of this. It is also clear that what the model learned was much more than just memorization. For instance, the model produced the perfectly valid name ``Michael Miller'', even though Michael and Miller never occurred together in the training set.  Additionally, it produced the name ``Connie Hifts'', which also sounds relatively plausible. The token ``Hifts'' is not in the training set, but happens to be a real (albeit rare) surname. In addition to names, some of the words the model produced never occurred in the training set. For instance, the word ``corporators'' is a valid English word. The model produced this word, even though there are no examples of ``corporators'' or ``corporator'' in the training set. Furthermore, the model properly used ``corporators'' like a noun (it followed ``of''). This ability to model plausible but rarely or never seen words is a potential advantage of subword models.

\section{Wikipedia experiments}

\subsection{Methods}
Another experiment was carried out on the 100MB subset of a Wikipedia corpus that is sometimes used as a test set for a larger Wikipedia corpus, and was originally used for a compression competition known as the Hutter prize \citep{Hutter-2006}. As compared with the Penn Treebank corpus, the Wikipedia corpus has far more unique words and more punctuation. It also includes some sections that are mostly XML, so the RNN must learn to model both English text and mark up language.

This set of experiments was designed to be set up in a similar way to a past study that used a 7-layer stacked LSTM with 21.3 million parameters \citep{Graves-2013}. This past study used the first 96 million characters for training, and the final 4 million characters for validation. This present experiment used an mLSTM with 21.2 million trainable parameters, using approximately 95 million characters for training, 1 million for validation, and 4 million for testing. The one slight difference between the dataset used here and the dataset in the past study is that in the present study around $0.4\%$ of the characters were lost from the dataset because they were not recognized by MATLAB. However, even if these unrecognised characters were encoded with a full byte each, the cross entropy error would only be increased by a maximum of about 0.03 bits per character. Because certain combinations of rare characters are often clumped together, it is likely that an RNN model could do much better than this. Two experiments were carried out, one using structural damping and another using line-search damping. For structural damping, $\mu$ was increased to 1 and Tikhonov damping with $\lambda=10$ was also used. The reason for this was to limit unstable updates.

The training set was broken into 2.1 million character subsets. For each training iteration, the gradient was computed on the full 2.1 million character subset and a conjugate gradient run was computed with a random set of one tenth of the subset's training examples. Batches were split into sequences of length 1000. The structural damping experiment was run on 4 Geforce GTX 970s (4GB) and the line-search damping experiment was run on 2 Geforce GTX 980s (4GB) and 1 Nvidia Tesla 40k (12GB). The GPUs were utilized using MATLAB's Parallel Computing Toolbox. To save memory on these GPUs, it was essential to use the hidden state recomputing strategy presented in Chapter 3.3.

\subsection{Results}

The results of the experiments are given in Table 5.2. These results indicate a very similar performance to the 7-layer stacked LSTM that had previously been trained on this dataset, which achieved a validation error of 1.67 bits/char \citep{Graves-2013}. This previous result can be considered the state of the art for RNNs with static weights, although the author was able to achieve a much better result with online weight adaptation that occurred during test time, which was not attempted in this experiment. There are several major differences in methodology between the past study and this present study. First of all the past study used minibatch stochastic gradient descent with derivative clipping as a learning algorithm, while this present study used Hessian free optimization. Hessian free optimization has the potential to be more computationally efficient, so it is encouraging that it was able to obtain a result competitive with a state of the art first order training method. Secondly, the past study used 7 layers of stacking to improve the expressiveness of LSTM, while this present study used no stacking, but had multiplicative connections. One can assume that the author of the past study believed that stacking was quite important for performing well on this dataset, as 7 layers is quite deep. So it is interesting to see here that a similar performance was achieved using no stacking, and only multiplicative connections to gain an advantage in expressiveness. It is possible that the utilization of both stacking and multiplicative connections could lead to further improvements on this dataset.

\begin{table}[h]
\begin{center}
\begin{tabular}{ | l | l | l | l | l |} \hline Architecture/Damping & Val Error & Test Error & Number of parameters \\  \hline mLSTM Structural & 1.70 & 1.71 & 21188000\\ \hline mLSTM line-search &  1.68 & 1.69 & 21188000\\ \hline
\end{tabular}
\end{center}
\caption{Results of multiplicative LSTM on Wikipedia corpus in bits/char}
\end{table}

\subsection{Samples from model}

As with the previous section, these models were used to generate samples of text. A 1000 character sample from each model with a beginning context of '$<$' is given below.

\hfill \break
Sample from structural damping model
\begin{spverbatim}
</comment>
      <text xml:space="preserve">:''This is a traveling influence in the [[Septactic Marinar]] and the G-zil tractaking the theory. Brooklines frequently seen as Captain Burrosa to build a chapter to the bowsler for the force.  He became an actor and liberal and critic, most Jewish and Ireland was openton dancing divisions.

Firstly, antenro-archs-hitter was one of ten years after the ground rat acoustilling. He served as [[Kashmira]] (quintel, Frianfoffek), and Caroline, England
[[Category:balance physics]]
**[[MPGAR]]
*[[Bible]]
*[[Audieracci]], a highly debate of the book ''Thame of Hat Earth''.

== Sports==
* ''[[The Mutanian arria]]'' &lt;br /&gt;&amp;nbsp;&amp;nbsp;&amp;nbsp;[[root and run]]&quot; or &quot;[[convention]]&quot; things &lt;BR&gt;
* [[English]] 60%
\end{spverbatim}
\hfill \break
Sample from line-search damping model

\begin{spverbatim}
</id>
      <timestamp>2006-02-05T03:46:44Z</timestamp>
      <contributor>
        <ip>20.95.287.168</ip>
      </contributor>
      <comment>/* Genus ===
Forty years painter, puff
**3/B.A. L.
*** Letters; George, Gerry F., page at [ram-gzwolystic gospel}}
{{See also===
* [[Guantangusian Falbure]]&lt;br&gt;
''Garfield Strateg&lt;/td&gt;
&lt;tr&gt;&lt;td&gt;1&lt;/td&gt;&lt;td&gt;1 v(1, 5 }''&quot;;)
# Was MIP (1), content with into the setting of complexes, see [[secrecies]] and [[length]] periods. 0-212 oth in Fantasy, the vulgarised region.

===October 1906:]]
This appear to refer to individual self-regulating physics behind the [[USA General Election, 1940|''USAL Precision or CHLIO]]'''&lt;br /&gt;[[M II Jickey]] (1975)
* [[Delano Paulina]] ([[Kathula]], [[South Carolina]], [[IUPAB]], [[OLGOR]] -- [[Winkleideses]]
|[[Lonson|Leslie]] | last = Phillips | publisher = S.E.  camp]]
[[Category:Ancient Greeks]]

[[bg:Gemu | jokuszkrj]]
[[tr:Grama]]
[[fr:Malap}}

After a week was surprised b
\end{spverbatim}

\hfill \break
For comparison, an excerpt from the real Wikipedia corpus is given below.

\begin{spverbatim}
[[Category:Demographics by country|Cyrpus]]
[[Category:Cypriot society]]

[[es:Demografía de Chipre]]</text>
    </revision>
  </page>
  <page>
    <title>Politics of Cyprus</title>
    <id>5597</id>
    <revision>
      <id>39374474</id>
      <timestamp>2006-02-12T19:20:22Z</timestamp>
      <contributor>
        <ip>83.121.2.141</ip>
      </contributor>
      <comment>disambiguation from [[FAO]] to [[Food and Agriculture Organization]] by the [[User:DabMachine|DabMachine]]</comment>
      <text xml:space="preserve">{{Politics of Cyprus}}
:''This entry is about '''politics of Cyprus''', especially the island of Cyprus and the Republic of Cyprus. For information on politics of Northern Cyprus, see the [[Politics of Northern Cyprus]].''
[[Cyprus]] is a divided island. Since [[1974]], the Greek Cypriot-led government (The [[Republic of Cyprus]]) has controlled the south two thirds, and the [[seperatist]] [[TRNC|Turkish Cypriot authorities]] the northern one-third. The Government of the [[Republic of Cyprus]] has continued as the sole internationally-recognized authority on the island, though in practice its power extends only to the Greek Cypriot-controlled area.

==United Cyprus==
The [[1960]] Cypriot [[Constitution]] provided for a [[presidential system]] of government with [[separation of powers|independent]] [[executive (government)|executive]], [[legislative]], and [[judicial]] branches, as well as a complex system of checks and balances including a weighted power-sharing ratio designed to protect the interests of the [[Turkish Cypriots]]. The executive, for example, was headed by a [[Greek Cypriot]] president and a Turkish Cypriot vice president, elected by their respective communities for five-year terms and each possessing a right of veto over certain types of [[legislation]] and executive decisions.

\end{spverbatim}

It is clear from these samples that the models learned many structural regularities in both XML and English. In one of its more impressive feats, the line-search damping model was able nest a realistic looking IP address inside a contributor tag with correct syntax, all spanning over about 50 characters. The text generated by the model is similar to but still distinguishable from text in the real corpus.

\subsection{Time-lag experiment}
It is difficult to estimate the full time-lag capabilities of a model simply by sampling from it. Time-lag capabilities on character level models are often assessed by the model's ability to close parentheses, braces, or brackets over a long time context. In this particular corpus, '[[' closed by ']]' is a common occurrence in the training set. If given the context of '[[', the model will almost always close with ']]' relatively quickly, but this does not mean it is not capable of using this context to influence its predictions for longer. As an experiment to measure time-lag capabilities of the model (model trained with structural damping was chosen arbitrarily), the model was given the initial context '[[' and sampled from. It was limited to only outputting alphabetic characters and spaces, but still predicted a probability distribution over all characters, with the idea that if the model is using its initial context, then the ratio of $\frac{P(']')}{P('[')}$ that the model predicts should be high, since in the training data a '[[' is almost always closed with a ']]'  before opening a new bracket. This ratio can be measured over time steps as the RNN generates more alphabetic text to see how long this context is being used. To avoid having to make assumptions about the baseline value of $\frac{P(']')}{P('[')}$, control experiments were also carried out with a beginning context of 'Th'. Ratios of $\frac{P(']')}{P('[')}$ according to the model were measured for 1000 time steps. A log (base 10) was applied to the ratio, and the data were smoothed by taking the means for sets of 10 time steps, leaving 100 data points. The data for the experimental context and the control context were then each averaged across 10 trials. The results in Figure 5.1 indicate that the ratio $\frac{P(']')}{P('[')}$ was much higher for the '[[' context than for the control context for the full 1000 time steps. This means that in certain situations, the model was capable of storing contextual information about characters atleast 1000 time steps in the past, and likely longer because the ratio showed no signs of decaying.

 \begin{figure}[h]
 \includegraphics[width=1.0\textwidth]{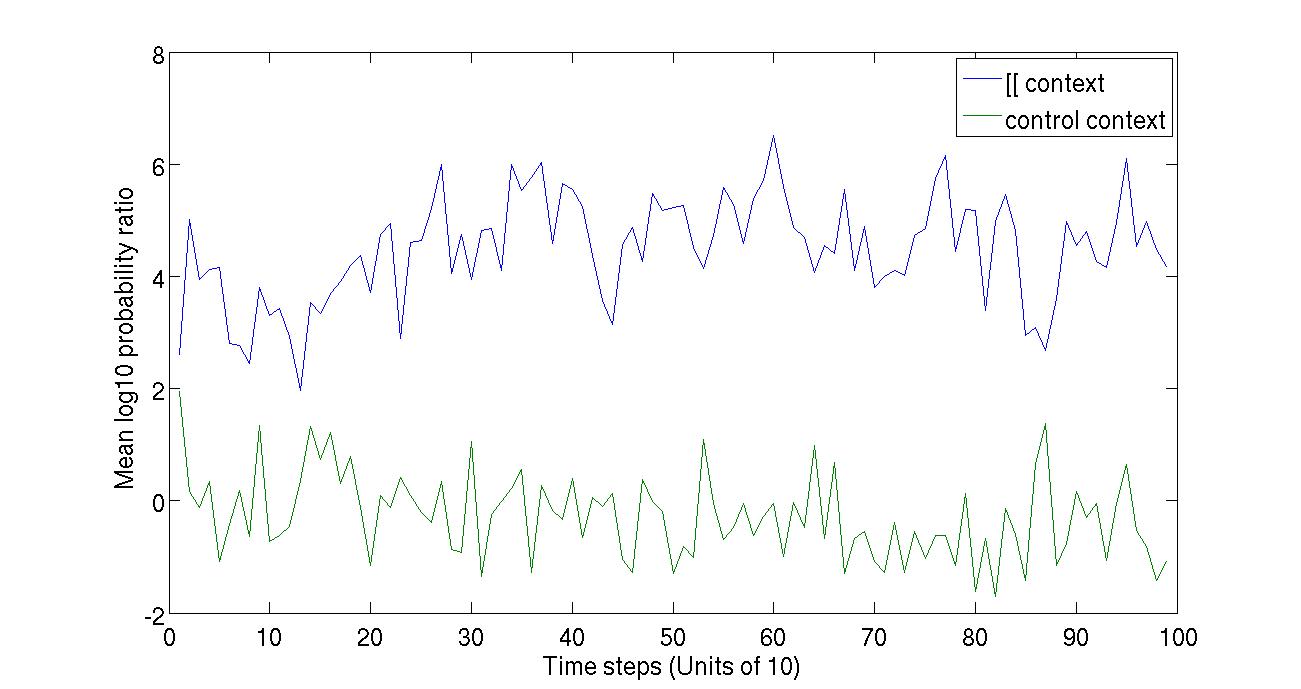}
\caption{Mean smoothed log ratio of $\frac{P(']')}{P('[')}$ given for the context of '[[', and control context of 'Th', for 1000 time steps after context was presented. Results are smoothed (log ratios for sets of 10 time steps were averaged), and averaged over 10 independent trials each. RNN was only allowed to output alphabetic characters and spaces, but still predicted probability distribution over all characters.}
\end{figure}

\chapter{Conclusion}

These experiments first demonstrated comparisons between several optimization methods and architectures on a smaller dataset. This included a comparison of damping methods, an LSTM trained with Hessian free optimization, a stacked mRNN variant, and a novel architecture that combined LSTM and multiplicative RNNs. The novel multiplicative LSTM hybrid resulted in improvements compared with both a multiplicative RNN and an LSTM alone. The multiplicative LSTM was then evaluated on two larger datasets. On the raw Penn Treebank corpus, over-fitting was a problem due to the small size of the corpus. Results were not directly comparable to published work with RNNs, that all used a preprocessed and simplified version of Penn Treebank. However, the model's success at modelling complex features in the raw corpus indicate that it likely would have been competitive with published work on the simpler preprocessed corpus. On the larger wikipedia corpus, the multiplicative LSTM model scored competitively with published results, using very different methodology. The model also demonstrated the ability to remember to close brackets over a time lag of atleast 1000 characters.

While Hessian free optimization has proven to be a reasonable way to train RNNs, it is far from being the full solution to the problem. For architectures such as multiplicative RNNs with especially high curvature, second order methods like Hessian free optimization are likely the most viable solution. However for general RNNs and LSTMs, Hessian free optimization is one of several possible solutions to the training difficulties. Hessian free optimization has the potential be faster than other methods for training RNNs on large datasets. Compared with first order methods, Hessian free optimization is able to better utilize parallelization, but also usually requires more total computation. Hessian free optimization tends to be most efficient when batch sizes are high, and the computations are split onto many processors. It is still unclear whether Hessian free optimization is able to generalize better than other learning algorithms, however, far more work has gone into finding regularization techniques for RNNs trained with stochastic gradient descent \citep{Zaremba-2014}.

These experiments demonstrated the expansion of Hessian free optimization to several new architectures and to larger RNNs. The results indicate that these methods are able to produce models competitive with state of the art RNN models in character prediction. Future work will be needed to develop regularization  methods that improve generalization of RNNs trained with this methodology.

\chapter{Appendix}

\section{LSTM R-forward pass}

\begin{equation}
R(H_{in}(t))= R(W_{hi})I(t) + R(W_{hh})H_{out}(t-1) + W_{hh}R(H_{out}(t-1));
\end{equation}

\begin{equation}
R(\omega(t))= R(W_{\omega i})I(t) + R(W_{\omega h})H_{out}(t-1) + W_{\omega h}R(H_{out}(t-1));
\end{equation}

\begin{equation}
R(\phi(t))= R(W_{\phi i})I(t) + R(W_{\phi h})H_{out}(t-1) + W_{\phi h}R(H_{out}(t-1));
\end{equation}

\begin{equation}
R(\rho (t))= R(W_{\rho i})I(t) + R(W_{\rho h})H_{out}(t-1) + W_{\rho h}R(H_{out}(t-1));
\end{equation}
\begin{equation}
\begin{split}
R(H_{state}(t)) = H_{in}(t) \odot R(\omega(t)) + R(H_{in}(t)) \odot \omega(t) \\ + R(\phi(t))\odot H_{state}(t-1) + \phi(t)\odot R(H_{state}(t-1))
\end{split}
\end{equation}

to simplify equations, define intermediate state $\zeta$ as input to $H_{out}$ before $\tanh$ is applied

\begin{equation}
R(\zeta(t)) = R(H_{state}(t)) \odot \rho(t) + H_{state}(t) \odot R(\rho(t)))
\end{equation}
\begin{equation}
R(H_{out}(t)) = R(\zeta(t)) \odot (1- H_{out}(t) \odot H_{out}(t))
\end{equation}
\begin{equation}
 R(O(t)) = W_{oh}R(H_{out}(t))+ R(W_{oh})(H_{out}(t))
\end{equation}
\section{LSTM backward pass}

When computing gradient, set  $\frac{\partial E}{\partial (O(t))} = O(t) - \gamma(t)$, when computing the R backward algorithm, set $\frac{\partial E}{\partial (O(t))} = H_{\sigma}(O(t))R(O(t))$

  \begin{equation}
  \frac{\partial E}{\partial W_{oh}} = \sum_{t=1}^T \frac{\partial E}{\partial O(t)} H_{out}(t)^T
   \end{equation}

\begin{equation}
\begin{split}
 \frac{\partial E}{\partial H_{out}(t)} =  W_{oh}^T \frac{\partial E}{\partial O(t)} + W_{hh}^T \frac{\partial E}{\partial H_{in}(t+1)} + W_{\omega h}^T \frac{\partial E}{\partial\omega_{in}(t+1)}+ \\ W_{\phi h}^T \frac{\partial E}{\partial \phi_{in}(t+1)} + W_{\rho h}^T \frac{\partial E}{\partial\rho_{in}(t+1)}
\end{split}
\end{equation}

to simplify equations, define intermediate state $\zeta$ as input to $H_{out}$ before $\tanh$ is applied
\begin{equation}
\frac{\partial E}{\partial\zeta(t)} = \frac{\partial E}{\partial H_{out}(t)} \odot (1-H_{out}(t) \odot H_{out}(t))
\end{equation}

 \begin{equation}
 \frac{\partial E}{\partial \rho_{out}(t)} = \frac{\partial E}{\partial\zeta(t)} \odot H_{state}(t)
  \end{equation}
  \begin{equation}
   \frac{\partial E}{\partial H_{state}(t)} = \frac{\partial E}{\partial\zeta(t)}\odot \rho(t) + \phi(t+1) \odot \frac{\partial E}{\partial H_{state}(t+1)}
  \end{equation}
  \begin{equation}
 \frac{\partial E}{\partial\phi_{out}(t)} = H_{state}(t-1) \odot \frac{\partial E}{H_{state}(t)}
  \end{equation}
    \begin{equation}
 \frac{\partial E}{\partial \omega_{out}(t)} =  H_{in}(t) \odot \frac{\partial E}{\partial H_{state}(t)}
  \end{equation}
  \begin{equation}
 \frac{\partial E}{\partial H_{in}(t)} =  \omega(t) \odot \frac{\partial E}{\partial H_{state}(t)}
  \end{equation}
  \begin{equation}
  \frac{\partial E}{\partial \omega_{in}(t)} = \frac{\partial E}{\partial \omega_{out}(t)} \odot \omega(t) \odot (1- \omega(t))
  \end{equation}

   \begin{equation}
  \frac{\partial E}{\partial\phi_{in}(t)} = \frac{\partial E}{\partial\phi_{out}(t)} \odot \phi(t) \odot (1- \phi(t))
  \end{equation}
     \begin{equation}
  \frac{\partial E}{\partial\rho_{in}(t)} = \frac{\partial E}{\partial\rho_{out}(t)} \odot \rho(t) \odot (1- \rho(t))
  \end{equation}
 \begin{equation}
  \frac{\partial E}{\partial W_{hh}} = \sum_{t=2}^T \frac{\partial E}{\partial H_{in}(t)}H_{out}(t-1)^T
   \end{equation}
 \begin{equation}
  \frac{\partial E}{\partial W_{\omega h}} = \sum_{t=2}^T \frac{\partial E}{\partial \omega_{in}(t)}H_{out}(t-1)^T
   \end{equation}
    \begin{equation}
  \frac{\partial E}{\partial W_{\phi h}}= \sum_{t=2}^T \frac{\partial E}{\partial \phi_{in}(t)}H_{out}(t-1)^T
   \end{equation}
    \begin{equation}
  \frac{\partial E}{\partial W_{\rho h}} = \sum_{t=2}^T \frac{\partial E}{\partial\rho_{in}(t)}H_{out}(t-1)^T
   \end{equation}

    \begin{equation}
  \frac{\partial E}{\partial W_{hi}} = \sum_{t=1}^T \frac{\partial E}{\partial H_{in}(t)}I(t)^T
   \end{equation}
 \begin{equation}
  \frac{\partial E}{\partial W_{\omega i}} = \sum_{t=1}^T \frac{\partial E}{\partial \omega_{in}(t)}I(t)^T
   \end{equation}
    \begin{equation}
  \frac{\partial E}{\partial W_{\phi i}} = \sum_{t=1}^T \frac{\partial E}{\partial \phi_{in}(t)}I(t)^T
   \end{equation}
    \begin{equation}
  \frac{\partial E}{\partial W_{\rho i}} = \sum_{t=1}^T \frac{\partial E}{\partial\rho_{in}(t)}I(t)^T
   \end{equation}

  \section{Multiplicative LSTM R-forward pass}

\begin{equation}
\begin{split}
R(M(t)) = R(W_{mi})I(t) \odot (W_{mh}H_{out}(t-1)) + \\ W_{mi}I(t) \odot(R(W_{mh})H_{out}(t-1) + W_{mh}R(H_{out}(t-1))
\end{split}
\end{equation}
\begin{equation}
R(H_{in}(t))= R(W_{hi})I(t) + R(W_{hm})M(t) + W_{hm}R(M(t));
\end{equation}

\begin{equation}
R(\omega(t))= R(W_{\omega i})I(t) + R(W_{\omega m})M(t) + W_{\omega m}R(M(t));
\end{equation}

\begin{equation}
R(\phi(t))= R(W_{\phi i})I(t) + R(W_{\phi m})M(t) + W_{\phi m}R(M(t));
\end{equation}

\begin{equation}
R(\rho (t))= R(W_{\rho i})I(t) + R(W_{\rho m})M(t) + W_{\rho m}R(M(t));
\end{equation}
\begin{equation}
\begin{split}
R(H_{state}(t)) = H_{in}(t) \odot R(\omega(t)) + R(H_{in}(t)) \odot \omega(t) \\ + R(\phi(t))\odot H_{state}(t-1) + \phi(t)\odot R(H_{state}(t-1))
\end{split}
\end{equation}

to simplify equations, define intermediate state $\zeta$ as input to $H_{out}$ before $\tanh$ is applied

\begin{equation}
R(\zeta(t)) = R(H_{state}(t)) \odot \rho(t) + H_{state}(t) \odot R(\rho(t)))
\end{equation}
\begin{equation}
R(H_{out}(t)) = R(\zeta(t)) \odot (1- H_{out}(t) \odot H_{out}(t))
\end{equation}

\begin{equation}
 R(O(t)) = W_{oh}R(H_{out}(t))+ R(W_{oh})(H_{out}(t))
\end{equation}

\section{Multiplicative LSTM backward pass}

When computing gradient, set  $\frac{\partial E}{\partial (O(t))} = O(t) - \gamma(t)$, when computing the R backward algorithm, set $\frac{\partial E}{\partial (O(t))} = H_{\sigma}(O(t))R(O(t))$

  \begin{equation}
  \frac{\partial E}{\partial W_{oh}} = \sum_{t=1}^T \frac{\partial E}{\partial O(t)}H_{out}(t)^T
   \end{equation}

\begin{equation}
\begin{split}
 \frac{\partial E}{\partial M(t)} =   W_{hm}^T \frac{\partial E}{\partial H_{in}(t)} + W_{\omega m}^T \frac{\partial E}{\partial\omega_{in}(t)}+ \\ W_{\phi m}^T \frac{\partial E}{\partial\phi_{in}(t)} + W_{\rho m}^T \frac{\partial E}{\partial\rho_{in}(t)}
\end{split}
\end{equation}

define $\chi$ as the intermediate state $W_{mi}I(t)$ and $\xi$ as the intermediate state $W_{mh}H(t-1)$

 \begin{equation}
 	\frac{\partial E}{\partial\chi(t)} = \frac{\partial E}{\partial M(t)} \odot \xi(t)
\end{equation}
\begin{equation}
	\frac{\partial E}{\partial \xi(t)} = \frac{\partial E}{\partial M(t)} \odot \chi(t)
\end{equation}
 \begin{equation}
  \frac{\partial E}{\partial W_{mh}} = \sum_{t=2}^T \frac{\partial E}{\partial \xi(t)}H_{out}(t-1)^T
   \end{equation}

    \begin{equation}
  \frac{\partial E}{\partial W_{mi}} = \sum_{t=1}^T \frac{\partial E}{\partial \chi(t)}I(t)^T
   \end{equation}

\begin{equation}
 \frac{\partial E}{\partial H_{out}(t)} = W_{mh}^T \frac{\partial E}{\partial \xi(t+1)} + W_{oh}^T \frac{\partial E}{\partial O(t)}
\end{equation}

to simplify equations, define intermediate state $\zeta$ as input to $H_{out}$ before $\tanh$ is applied
\begin{equation}
\frac{\partial E}{\partial \zeta(t)} = \frac{\partial E}{\partial H_{out}(t)} \odot (1-H_{out}(t) \odot H_{out}(t))
\end{equation}

 \begin{equation}
 \frac{\partial E}{\partial \rho(t)_{out}} = \frac{\partial E}{\partial \zeta(t)} \odot H_{state}(t)
  \end{equation}
  \begin{equation}
   \frac{\partial E}{\partial H_{state}(t)} = \frac{\partial E}{\partial\zeta(t)}\odot \rho(t) + \phi(t+1) \odot \frac{\partial E}{\partial H_{state}(t+1)}
  \end{equation}
  \begin{equation}
 \frac{\partial E}{\partial\phi_{out}(t)} = H_{state}(t-1) \odot \frac{\partial E}{\partial H_{state}(t)}
  \end{equation}
    \begin{equation}
 \frac{\partial E}{\partial\omega_{out}(t)} =  H_{in}(t) \odot \frac{\partial E}{\partial H_{state}(t)}
  \end{equation}
  \begin{equation}
 \frac{\partial E}{\partial H_{in}(t)} =  \omega(t) \odot \frac{\partial E}{\partial H_{state}(t)}
  \end{equation}
  \begin{equation}
  \frac{\partial E}{\partial \omega_{in}(t)} = \frac{\partial E}{\partial \omega_{out}(t)} \odot \omega(t) \odot (1- \omega(t))
  \end{equation}

   \begin{equation}
  \frac{\partial E}{\partial\phi_{in}(t)} = \frac{\partial E}{\partial\phi_{out}(t)} \odot \phi(t) \odot (1- \phi(t))
  \end{equation}
     \begin{equation}
  \frac{\partial E}{\partial\rho_{in}(t)} = \frac{\partial E}{\partial\rho_{out}(t)} \odot \rho(t) \odot (1- \rho(t))
  \end{equation}
 \begin{equation}
  \frac{\partial E}{\partial W_{hm}} = \sum_{t=1}^T \frac{\partial E}{\partial H_{in}(t)}M(t)^T
   \end{equation}
 \begin{equation}
  \frac{\partial E}{\partial W_{\omega m}} = \sum_{t=1}^T \frac{\partial E}{\partial \omega_{in}(t)}M(t)^T
   \end{equation}
    \begin{equation}
  \frac{\partial E}{\partial W_{\phi m}} = \sum_{t=1}^T \frac{\partial E}{\partial \phi_{in}(t)}M(t)^T
   \end{equation}
    \begin{equation}
  \frac{\partial E}{\partial W_{\rho m}} = \sum_{t=1}^T \frac{\partial E}{\partial \rho_{in}(t)}M(t)^T
   \end{equation}

    \begin{equation}
  \frac{\partial E}{\partial W_{hi}} = \sum_{t=1}^T \frac{\partial E}{\partial H_{in}(t)}I(t)^T
   \end{equation}
 \begin{equation}
  \frac{\partial E}{\partial W_{\omega i}} = \sum_{t=1}^T \frac{\partial E}{\partial \omega_{in}(t)}I(t)^T
   \end{equation}
    \begin{equation}
  \frac{\partial E}{\partial W_{\phi i}} = \sum_{t=1}^T \frac{\partial E}{\partial \phi_{in}(t)}I(t)^T
   \end{equation}
    \begin{equation}
  \frac{\partial E}{\partial W_{\rho i}} = \sum_{t=1}^T \frac{\partial E}{\partial \rho_{in}(t)}I(t)^T
   \end{equation}

\bibliographystyle{apalike}

\bibliography{thesis}

\end{document}